\documentclass[Afour,sageh,times]{sagej}

\usepackage{moreverb,url}

\usepackage[colorlinks,bookmarksopen,bookmarksnumbered,citecolor=blue,urlcolor=red]{hyperref}

\usepackage{graphicx} 
\usepackage{amsmath} 
\usepackage{amssymb}  

\usepackage{lipsum}
\usepackage{color}
\usepackage{cite}
\usepackage{algpseudocode}
\usepackage[ruled]{algorithm}

\usepackage{diagbox} 
\usepackage{tabu}
\usepackage{array}
\usepackage{multirow}
\usepackage{booktabs}
\usepackage[switch]{lineno}
\newcolumntype{P}[1]{>{\centering\arraybackslash}p{#1}}
\newcolumntype{M}[1]{>{\centering\arraybackslash}m{#1}}

\newcommand{\revision}[1]{\textcolor{black}{#1}} 
\newcommand{\final}[1]{\textcolor{black}{#1}} 

\newcommand\BibTeX{{\rmfamily B\kern-.05em \textsc{i\kern-.025em b}\kern-.08em
T\kern-.1667em\lower.7ex\hbox{E}\kern-.125emX}}

\def\volumeyear{2016}

\begin{document}

\runninghead{Jiang~\emph{et al.}}

\title{Intelligent Robotic Sonographer: Mutual Information-based Disentangled Reward Learning from Few Demonstrations}

\author{Zhongliang Jiang*$^{1}$, Yuan Bi*$^{1}$, Mingchuan Zhou$^{1}$, Ying Hu$^{2}$, Michael Burke$^{3}$ and Nassir Navab$^{1,4}$}

\affiliation{
$^{*}$ Authors are with equal contributions\\
\affilnum{1}The Chair for Computer Aided Medical Procedures and Augmented Reality, Technical University of Munich, Germany.\\
\affilnum{2}Shenzhen Institute of Advanced Technology, Chinese Academy of Science, China.\\
\affilnum{3}The Department of Electrical and Computer Systems Engineering, Monash University, Australia.\\
\revision{\affilnum{4}The Laboratory for Computer Aided Medical Procedures, Johns Hopkins University, Baltimore, MD 21218 USA.\\}
}

\corrauth{Zhongliang Jiang, the Chair for Computer Aided Medical Procedures and Augmented Reality, Technical University of Munich, Boltzmannstr. 3, 85748 Garching bei M\"unchen, Germany.}

\email{zl.jiang@tum.de}

\begin{abstract}
Ultrasound (US) imaging is widely used for biometric measurement and diagnosis of internal organs due to the advantages of being real-time and radiation-free. However, due to inter-operator variations, resulting images highly depend on the experience of sonographers. This work proposes an intelligent robotic sonographer to autonomously ``explore" target anatomies and navigate a US probe to a relevant 2D plane by learning from the expert. \final{The underlying high-level physiological knowledge from experts is inferred by a neural reward function, using a ranked pairwise image comparisons approach in a self-supervised fashion. This process can be referred to as understanding the ``language of sonography".} Considering the generalization capability to overcome inter-patient variations, mutual information is estimated by a network to explicitly disentangle the task-related and domain features in latent space.
The robotic localization is carried out in coarse-to-fine mode based on the predicted reward associated with B-mode images. 
\revision{To validate the effectiveness of the proposed reward inference network, representative experiments were performed on vascular phantoms (``line" target), two types of ex-vivo animal organs (chicken heart and lamb kidney) phantoms (``point" target) and in-vivo human carotids, respectively. To further validate the performance of the autonomous acquisition framework, physical robotic acquisitions were performed on three phantoms (vascular, chicken heart, and lamb kidney). }
The results demonstrated that the proposed advanced framework can robustly work on a variety of seen and unseen phantoms as well as in-vivo human carotid data.
\textbf{Code:} \url{https://github.com/yuan-12138/MI-GPSR}
\end{abstract}


\maketitle

\section{Introduction}
\begin{figure}[ht!]
\centering
\includegraphics[width=0.40\textwidth]{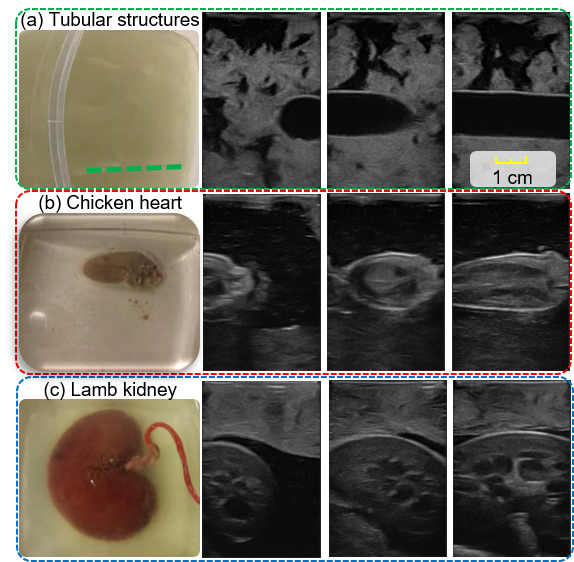}
\caption{Illustrations of standard planes for different organs. (a) the longitudinal view of a mimicked vesicular phantom, (b) the longitudinal view of a chicken heart, and (c) the longitudinal view of a lamb kidney. From the 2nd column to the 4th column are B-mode images of different objects recorded by a human operator manually aligning the probe pose towards standard planes. All the displayed US images share the same scale bar. US imaging depth is $5~cm$.
}
\label{Fig_standard_plane}
\end{figure}

Ultrasound (US) is one of the most widely used imaging techniques to visualize internal anatomies. Unlike computed tomography (CT) and magnetic resonance imaging (MRI), US imaging is real-time, low-cost, and radiation-free~\citep{hoskins2019diagnostic}. \revision{In conventional US examinations, sonographers often need to search for some standard US planes for diagnosis.
The standard US planes are important for performing quantitative biometric measurements~\citep{rizi2020carotid, baumgartner2017sononet}, image-guided interventions~\citep{stone2010needle, chen2020deep}, and identifying abnormalities, e.g., the longitudinal view of vascular structures (see Fig.~\ref{Fig_standard_plane}), which is often required to estimate the flow velocity in clinical practice~\citep{bi2022vesnet}.}
However, substantial experience and visuo-tactile skills are needed to correctly and accurately identify such planes from speckled US images. The potential inter- and intra-operator variations will challenge the achievement of consistent and repeatable diagnoses, particularly for novice sonographers. 
Therefore, the development of the robotic US system (RUSS), which can learn the underlying physiological knowledge from experienced sonographers to identify the standard US planes, is meaningful for obtaining standardized and operator-independent US images. This can also release sonographers from repetitive and cumbersome operations and reduce work-related musculoskeletal disorders.

Due to the advantages of accuracy, stability, and repeatability, robot-assisted US imaging has been investigated for decades. Pierrot~\emph{et al.} developed a RUSS that can maintain a constant force between the probe and patients for cardiovascular disease prevention~\citep{pierrot1999hippocrate}. Gilbertson~\emph{et al.} designed one degree of freedom (DoF) hand-held device to reduce force-induced image quality variation for free-hand scans~\citep{gilbertson2015force}. Conti~\emph{et al.} proposed a collaborative RUSS, which allows operators to instinctively operate a US probe remotely, while contact force is automatically maintained using compliant force control~\citep{conti2014interface}. To accurately control the US acquisition parameters (e.g., contact force and probe orientation) for stable and repeatable US images, Jiang~\emph{et al.} proposed a method to automatically position a linear probe to the normal direction of the contact surface using a confidence map and the estimated Cartesian force at the tool center point (TCP)~\citep{jiang2020automatic}. Then, a mechanical model-based method was proposed to adjust the orientation of both convex and linear probes~\citep{jiang2020automatic2}. In addition, Huang~\emph{et al.} and Chatelain~\emph{et al.} computed the desired probe orientation based on an RGB-D camera~\citep{huang2018robotic} and a visual servoing framework~\citep{chatelain2017confidence}, respectively. 
\final{
To estimate the region of interest, which refers to high-quality regions for examinations, Goel~\emph{et al.}~\citep{goel2022autonomous} and Raina~\emph{et al.}~\citep{raina2023deep,raina2023robotic} delved into the application of Bayesian Optimization methods in the context of RUSS.
}
The aforementioned RUSSs are promising to overcome the limitations of free-hand US, while the ability to learn from human experts to search for the standard US planes has not been fully researched yet.

\par
To develop an intelligent RUSS that can automatically identify and navigate the probe to the standard scan planes, various advanced learning-based approaches have been developed to improve the understanding of both images and the dynamic environment. Regarding carotid examination, Huang~\emph{et al.} divided the robot-assisted scanning procedure into three sub-stages to mimic the behavior of clinicians~\citep{huang2021towards}. 
Baumgartner~\emph{et al.} proposed the SonoNet to automatically detect fetal standard views during free-hand scans in real-time using convolutional neural networks~\citep{baumgartner2017sononet}.
To further consider automatically navigating a probe to the target US plane, Droste~\emph{et al.} trained the US-GuideNet to predict the final goal pose and the next movement action (rotation) based on the recorded US sweeps from experts~\citep{droste2020automatic}. Although the US-GuideNet was only validated in a virtual environment (recorded images from experts), its results proved the feasibility of learning such challenging tasks and understanding the human operator's intention from their demonstrations.
To automatically locate the sacrum in US view, Hase~\emph{et al.} trained a deep Q-network (DQN) to move the US probe in 2-DoF translational motion~\citep{hase2020ultrasound}. 
To fully control the 6-DoF motion of the probe, Li~\emph{et al.} presented a deep Reinforcement Learning (RL) framework to control the 6D pose of a virtual probe to a standard plane in simulation~\citep{li2021autonomous}. The results suggested that the proposed method can work effectively on intra-patient data ($92\%$), while the performance on unseen data ($46\%$) still has space for improvement.
The significant difference in the performance between trained and unseen data is caused by the variations in patients' data. In addition, the inherited characteristics of RL methods, like requiring rich dynamic contact with the environment and performance decrease in varied settings, will limit the deployment of RL-based approaches in real scenarios, particularly for clinical applications where patient safety is of the utmost importance.

\par
\final{In this work, we propose a novel machine learning framework to learn the underlying physiological knowledge from a few expert demonstrations and automatically navigate a probe to the desired US planes.} To guarantee the generalization capability of the proposed approach to unseen patients, mutual information is employed to explicitly disentangle the task-related and domain features for the input images in the latent space. To eliminate the need for cumbersome annotation, spatial ranking approaches are used to enable self-supervised training. After training, an estimated reward is computed for each individual image in real-time, which reflects the preference of the sonographers during the US examination. The main contributions are summarized as follows:
\begin{itemize}
  \item To understand expert sonographers' semantic reasoning and intention from a few demonstrations, a novel learning-based framework is proposed to predict the reward associated with the US images. The use of the probability ranking approach enables the training process to be done in the self-supervised mode without any requirement for cumbersome annotation.
  \item To deal with the two most common US scanning objectives (``point" and ``line" task) in one framework, we propose a global probabilistic spatial ranking (GPSR) network to generate unbiased image comparisons from all expert demonstrations. The GPSR can improve data efficiency, and it is especially valuable for sub-optimal expert demonstrations, where the target US planes are often observed multiple times during a single acquisition.
  \item To ensure the generalization capability, we integrate the Mutual Information~\citep{belghazi2018mutual} (MI) into GPSR (MI-GPSR) to explicitly disentangle the task-related features from the domain features. The MI-GPSR is validated to be able to properly estimate the rewards for US images on unseen phantoms and volunteers with significantly different geometrical sizes and image artifacts from the training data.
  \item To enable the possibility of applying robotic navigation in a real scenario, we propose a coarse-to-fine navigation framework instead of using RL or informative path planning approaches. Considering the time efficiency, a 3D reconstruction volume is computed using a US scan of the target anatomy. Therefore, a large amount of US images is simulated from this 3D US volume and the coarse acquisition pose for the standard plane is obtained by finding the simulated image with maximum reward. Then a fine-tuning process is carried out around the coarse location to bridge the potential gap between simulated and real US images. 
\end{itemize}
Finally, to validate the performance of the proposed approach, experiments have been carried out in a simulated grid world and two representative cases on gel vascular phantoms, alongside more challenging ex-vivo animal organ phantoms (lamb kidney and chicken heart). \revision{To ensure transferability for real human tissues, the reward inference network was further validated on real data of volunteers' carotids.} The results demonstrate that the proposed MI-GPSR can properly predict the reward of US images \revision{from} unseen demonstrations of \revision{unseen phantoms and volunteers}.

\par
The remainder of this paper is organized as follows. Section II presents related work. The proposed MI-GPSR is presented in Section III. Section IV describes the details of the autonomous US-guided navigation procedures. The experimental results in grid-world, three representative types of phantoms (vascular, chicken heart, and lamb kidney phantoms), and \revision{real carotid data from volunteers} are described in Section V. The discussion about the current challenges, and potential directions are presented in Section VI. Finally, in section VII, the paper is concluded with a summary of the presented approach. 

\section{Related Work}
\subsection{Robotic US System}
\par
Due to high inter-operator variability and the lack of reproducibility of free-hand US examination, RUSS has been considered as a promising solution to achieve standardized acquisition and diagnosis.
Virga~\emph{et al.} presented an approach to automatically acquire 3D US images for abdominal aortic aneurysms screening~\citep{virga2016automatic}. The scan path was determined based on the registration between the patient's surface and a generic MRI-based atlas. \revision{To consider the non-rigid motion between human bodies, Jiang~\emph{et al.} proposed an atlas-based framework allowing autonomous robotic US screening for limb arteries by using a non-rigid registration between the arm surfaces point clouds~\citep{jiang2022towards}.} To eliminate the requirement for the pre-operative images, \revision{Jiang~\emph{et al.} developed an end-to-end workflow for autonomous robotic screening of tubular structures based only on real-time B-mode images~\citep{jiang2021autonomous}.} Huang~\emph{et al.} proposed a camera-based method to automatically determine the scan path~\citep{huang2018robotic}. To utilize human experience, Abolmaesumi~\emph{et al.} developed a shared control method to visualize carotid arteries in 3D~\citep{abolmaesumi2002image}. In addition, RUSS has been widely used in different applications, such as visualizing underlying bone surfaces~\citep{jiang2020automatic2} and imaging human spines~\citep{zhang2021self}. The aforementioned RUSSs are mainly developed to automatically and accurately visualize the anatomies of interest. However, the challenge of guiding a US probe to desired locations, displaying standard scan planes, is not considered in these approaches.

\par
\revision{Considering pandemics such as Covid-19, Akbari~\emph{et al.} developed a RUSS that allows the physical separation of sonographers and patients~\citep{akbari2021robotic}. Antico~\emph{et al.} employed the Bayesian CNN to segment the femoral cartilage and use the volumetric US as the guidance in minimally invasive robotic surgery for the knee~\citep{antico2020bayesian}. Kim~\emph{et al.} considered the real-time contact force feedback to perform a heart scan autonomously~\citep{kim2020robot}. Naidu~\emph{et al.} incorporated tactile sensing to augment US images to enhance the location accuracy of tumor tissues~\citep{naidu2017breakthrough}. Besides, Monfaredi~\emph{et al.} discussed the development of a parallel telerobotic system~\citep{monfaredi2015robot}. Due to the focus of this article, we cannot provide a comprehensive overview of RUSS. For readers who are interested in a deep overview of the field, there are several survey articles that have investigated the developments of RUSS from distinct perspectives. Von Haxthausen~\emph{et al.} systematically summarized the related publications between 2016 and 2020~\citep{von2021medical}. Li~\emph{et al.} categorized the existing RUSS in terms of the level of automation~\citep{li2021overview}. In addition, Jiang~\emph{et al.} provided a comprehensive survey including both teleoperated and autonomous RUSS, which also systematically summarized the related technologies, such as enabling techniques, advanced application-oriented techniques, and emerging learning-based approaches~\citep{jiang2023robotic}.}

\subsection{Detection and Navigation of Standard Planes}
\par
Due to the potential deformation of soft tissue and the often hard-to-interpret US imaging, guiding a probe to correct planes is a highly sophisticated task, which requires years of training~\citep{maraci2014searching}. In addition, limited by human hand-eye coordination ability, these tasks suffer from low reproducibility and large intra-operator variations~\citep{chan2009volumetric}. These drawbacks severely limit the clinical acceptance of US modality for tasks requiring repeatable, quantitative, and accurate measurements, i.e., monitoring tumor growth. To address this challenge, Chen~\emph{et al.} proposed a learning-based approach to locate the fetal abdominal standard plane in US videos using a deep convolutional neural network (CNN)~\citep{chen2015standard}. Each input video frame was processed by a classifier to detect the standard plane using a sliding window. 
To further eliminate the requirement for the cumbersome annotation, Baumgartner~\emph{et al.} proposed the SonoNet using weak supervision to detect standard views during US scans in real-time~\citep{baumgartner2017sononet}. The experiments validated that the SonoNet can effectively extract multiple standard planes on free-hand US images.

\par
To automatically navigate a probe to standard planes, Droste~\emph{et al.} trained a policy network to estimate the next probe movement aiming to mimic the expert behavior~\citep{droste2020automatic}. The network included the gated recurrent unit (GRU) and was trained based on the recorded consecutive US images and probe orientations extracted from expert demonstrations. Due to the limitation of the inertial measurement unit (IMU), this work was only validated in terms of rotational movement, while there is no technical gap in applying the proposed method to the translational motion.
Hase~\emph{et al.} trained a DQN to navigate a probe to sacrum in a virtual 2D grid word, and a binary classifier was used to terminate the RL-based searching process~\citep{hase2020ultrasound}. To navigate to the paramedian sagittal oblique plane (a standard plane used in spine US examination), Li~\emph{et al.} considered both rotational and translational movements in a simulated environment using a deep RL framework~\citep{li2021autonomous}. This approach can effectively work on the images recorded from patients included in the training dataset ($92\%$), while further development is needed to improve the generalisability on unseen patients ($46\%$). 
\revision{To autonomously navigate a probe to the longitudinal view of tubular structures, Bi~\emph{et al.} proposed a method with high generalization capability by using segmented binary masks as the inputted state to an RL agent~\citep{bi2022vesnet}. The use of binary masks bridges the gap between the simulation and B-mode images obtained in real scenarios. Without any fine-tuning steps, the trained VesNet-RL model can reliably navigate the probe to the standard plane on a vascular phantom ($92.3\%$) and  on in-vivo carotid volumes from a volunteer ($91.5\%$).}


\subsection{Learning from Demonstration}
\par
Since RL algorithms often require rich interaction with the environment to learn the policies for a given task~\citep{kurenkov2020ac}, most of the aforementioned studies were trained in simulation or using recorded US examination videos rather than on real patients. Besides RL methods, some existing approaches tried to directly teach RUSS to perform US scans based on expert demonstrations, which have the potential to effectively alleviate the complexity of robotic programming. In addition, such a system can learn the underlying physiological knowledge from experienced experts and used the knowledge to help train young sonographers.
Learning from demonstration (LfD) can be broadly categorized into two approaches: imitation learning (IL)~\citep{ross2011reduction} and inverse reinforcement learning (IRL)~\citep{abbeel2004apprenticeship}. Imitation learning directly generates a predictive model to estimate the next action based on the current state, which requires optimal demonstrations because the fundamental logic of such approaches is to imitate the behavior rather than to understand the latent objective. On the contrary, IRL aims to represent a given task by a reward function with respect to state features, which often assumes that the reward value can be linearly or exponentially linked to the number of times of the features witnessed~\citep{abbeel2004apprenticeship} or the likelihood of features observed in a demonstration~\citep{ziebart2008maximum}. In other words, a higher reward should be assigned to the states more often appears. \final{Due to the acoustic shadows, poor contrast, speckle noise, and potential deformation in resulting images~\citep{mishra2018ultrasound}, guiding a probe to the standard US planes is sophisticated, even for senior sonographers. This means the experts' demonstrations of searching for standard US planes will be sub-optimal and even contradictory~\citep{burke2023learning}.} Therefore, the popular maximum-entropy IRL method~\citep{aghasadeghi2011maximum} cannot be directly applied in our applications.


\par
Specific to the field of robotic US, Mylonas~\emph{et al.} employed Gaussian Mixture Modeling (GMM) to model the demonstration in a probabilistic manner~\citep{mylonas2013autonomous}. This pioneering try only considered the trajectory rather than US images. 
In order to achieve good performance, it has strict requirements for the initial position and phantom position. Besides, Burke~\emph{et al.} introduced a probabilistic temporal ranking model which assumes that the images shown in the later stage are more important than the earlier images~\citep{burke2023learning}, allowing for reward inference from sub-optimal scanning demonstrations. They use this model to coarsely navigate a US probe to a mimicked tumor inside of a gel phantom, followed by an exploratory Bayesian optimization policy to search for scanning positions that capture images with high rewards. 

\subsection{Mutual Information-based Feature Disentanglement}
\par
Recent developments of deep neural networks (DNNs) demonstrated expert-level accuracy in a broad range of computer vision tasks like image segmentation~\citep{minaee2021image}, object detection~\citep{liu2020deep} and classification\citep{russakovsky2015imagenet}. However, the performance is often reduced when a trained model is applied to unseen domains, e.g., different scanner vendors and image acquisition settings. Considering the task of learning from clinical experts' demonstrations, the limited training datasets will result in poor generalization. Inspired by the way of feature representation in DNNs that progressively leads to more abstract features at deeper layers of DNNs, a feature disentanglement process is carried out to explicitly separate the task-related and domain features to improve the generalization capability on unseen data. Theoretically, explicit feature representations will have the potential to deal with previously unseen domains, which is particularly valuable for US applications.

\begin{figure*}[ht!]
\centering
\includegraphics[width=0.80\textwidth]{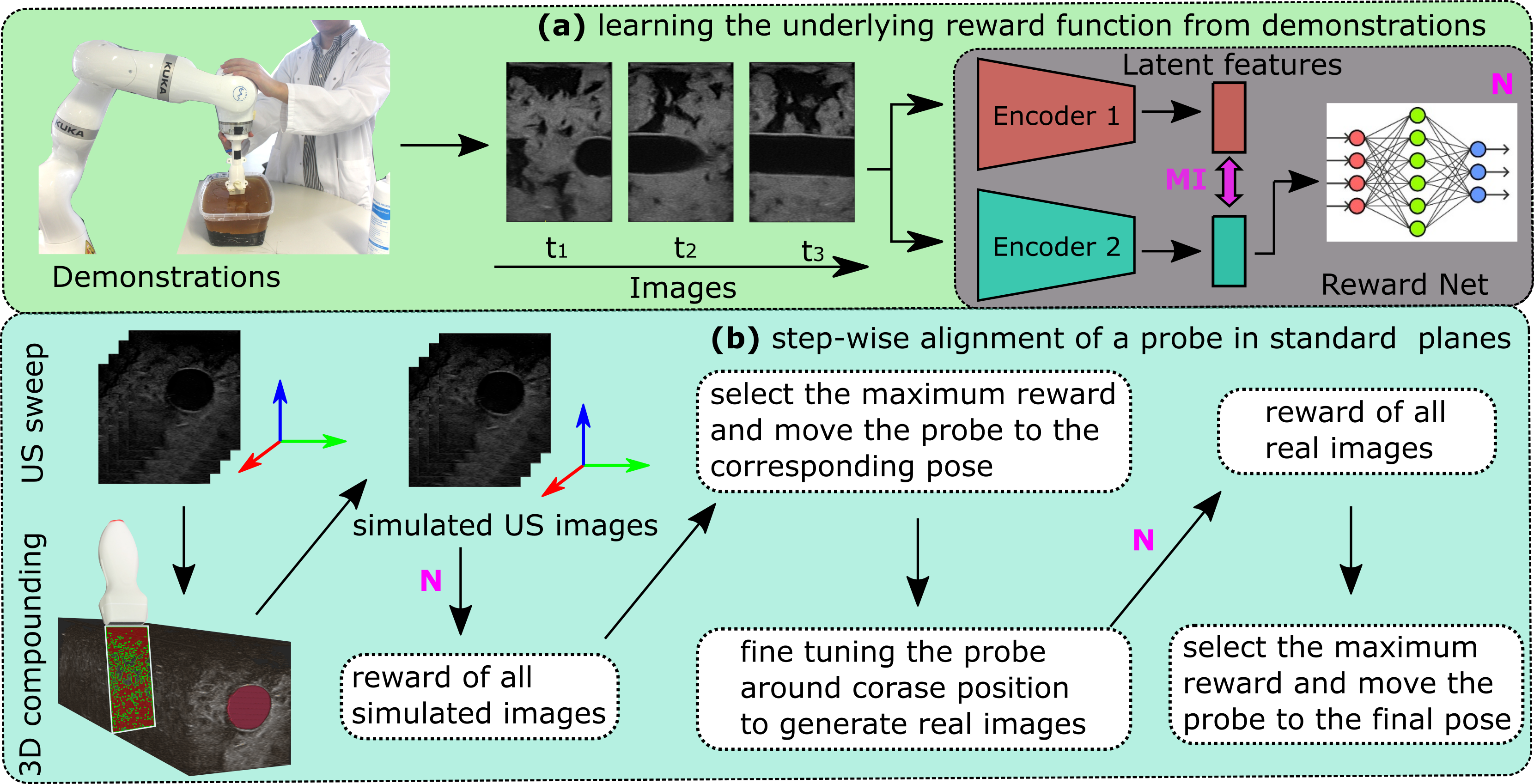}
\caption{Schematic overview of the proposed framework.}
\label{Fig_overview}
\end{figure*}

\par
\final{Mutual information (MI) is a fundamental quantification for computing the correlation between variables, which captures non-linear statistical dependencies. MI will be maximized when the two inputs share the same information. Therefore, to explicitly disentangle task-related and domain features, we can minimize the MI between these two feature representations of the same input in a deep layer~\citep{liu2021mutual}.} Nevertheless, the fast computation of MI is restricted to discrete variables~\citep{paninski2003estimation}. Due to the use of both marginal probability and joint probability of two variables, the computation complexity is related to the sample number. To achieve the computation of MI between high dimensional continuous random variables, Belghazi~\emph{et al.} proposed an MI Neural Estimator (MINE) by gradient descent over neural networks~\citep{belghazi2018mutual}. The MINE is scalable in dimension and sample size, and it provides unbiased MI estimations as well. Based on the MINE, Meng~\emph{et al.} developed MI-based Disentangled Neural Networks to extract generalizable categorical features from US images to transfer knowledge to unseen categories~\citep{meng2020mutual}. Based on the experiments, they claimed that the proposed method outperformed the state-of-the-art on the task of classification of unseen categories. Similarly, Liu~\emph{et al.} employed MI to extract discriminative representations to improve the generalization of recognition tasks against the disturbances of environment~\citep{liu2021mutual}. \revision{
Bi~\emph{et al.} use MI to explicitly disentangle the domain and anatomy features to enhance the generalization capability for US segmentation tasks~\citep{bi2023mi}.
}

\subsection{Proposed Approach}
\par
To assist the sonographer in achieving consistent and repeatable diagnosis during US examinations, we developed an intelligent RUSS to navigate the probe to standard scan planes by learning from few demonstrations. To the best of our knowledge, this is the first work that aims to learn the underlying skill and anatomical knowledge directly from human experts to align a probe to standard scan planes automatically. The ability to learn from demonstrations allows sonographers to intuitively transfer their physiological knowledge to a RUSS without the requirement of any robotic programming. This is still a challenging task in the area of robotic learning, particularly when the demonstrations are not optimal~\citep{eteke2020reward}. Inspired by~\citep{burke2023learning}, a global probabilistic spatial ranking (GPSR) method was developed to learn the latent skills and overcome the potentially biased results caused by inconsistent demonstrations. To increase the generalization capability and adapt the variation between patients, the MINE is employed (MI-GPSR) to explicitly disentangle the task-related features from the domain features.
The MI-GPSR mainly consists of two feature encoders to extract the task-related and domain features of the input images, a decoder to reconstruct the input images based on the extracted features in latent space, and a fully connected network (FCN) used to predict the reward for individual images. Considering the time efficiency, and the safety concerns about the rich interactions with patients, it is impractical to train an agent using RL~\citep{bi2022vesnet, li2021autonomous} or apply Gaussian process path planning~\citep{burke2023learning} in a real scenario. Therefore, we divided the navigation process into coarse and fine-tuning procedures. Regarding the first step, a US sweep over the target anatomies was performed, and then a virtual probe was generated on the upper surface of the compounded 3D volume. By assigning various probe poses, a larger number of simulated 2D images were created. Based on the predicted reward for individual synthetic images, a greedy algorithm is employed to detect the desired plane. 
Finally, to further bridge the gap between simulated images and real images, the probe was finely adjusted around the coarse location to achieve a more precise result, like human operators. An overview of the proposed approach is shown in Fig.~\ref{Fig_overview}. 

\par
In addition, to demonstrate the potential of the proposed approach, besides grid world analysis, the proposed method was validated on gel vascular phantoms and more challenging ex-vivo animal \revision{organ phantoms} (chicken heart and lamb kidney). In contrast to simulation, such tasks are more realistic and more challenging in a physical environment because the resulting images are sensitive to practical factors, i.e., probe pose and contact force. To further validate the generalization capability of the proposed MI-GPSR, the trained models are validated on the demonstrations acquired from unseen phantoms with different anatomical structures and image styles from the trained phantoms as well as unseen in-vivo demonstrations from unseen volunteers.

   


\section{Learning from Few Demonstrations}
\par
This section describes the probabilistic temporal ranking approach (PTR) and the proposed GPSR and MI-GPSR methods to learn the latent reward function for individual B-mode images from expert demonstrations. In addition, a confidence-based approach is developed to filter out the abnormal demonstrations to avoid inconsistent target standard US planes. \revision{It is noteworthy that the presented three approaches (PTR, GPSR, and MI-GPSR) are independent and can be used individually.}

\subsection{Probabilistic Temporal Ranking (PTR)} ~\label{sec:III-PTR}
Probabilistic Temporal Ranking (PTR) allows for uncertainty quantification and reward inference from demonstrations~\citep{burke2023learning}. Considering the characteristic of US examinations, experts often need to move a probe around to search for the target anatomy and finally stop at some specific US planes to do biometric measurements. This characteristic will result in sub-optimal US demonstrations, even for experienced sonographers. Therefore, the existing IRL approaches~\citep{abbeel2004apprenticeship,ziebart2008maximum} cannot be directly applied because the states or images observed most often in the sub-optimal demonstrations are not the states that should be assigned the highest reward. To tackle this problem, PTR incorporates an additional assumption that the images observed in the later stage of the demonstrated US sweeps should have a larger reward than the ones seen at an earlier stage. Based on this assumption, PTR is trained to infer the reward of individual images based on the pairwise comparisons sampled from expert demonstrations.

\par
For PTR, a convolutional variational autoencoder (CVAE) was used to extract the latent features ($z_t$) of each observation ($x_t$). The reward is predicted using an FCN $r_{\psi}(z_t)$ because an FCN is often used to approximate the Gaussian processes~\citep{neal1996priors}. To train the PTR, a differentiable neural approximation is employed with two observation inputs, i.e., pairwise image comparisons generated from demonstrations. To optimize the reward network parameters, the comparison outcome $g_t\in\{0, 1\}$ is used, which is automatically generated with respect to the temporal information of each frame in the demonstrations. Based on the assumption, ground truth of comparison outcome is $g_t=0$ if $t_1 > t_2$, and $g_t=1$ if $t_1 < t_2$. To use this self-supervised signal to train the PTR, the generative process for a pairwise comparison outcome is modeled as a Bernoulli trial as Eq.~(\ref{eq_Bernoulli}).
\begin{equation}\label{eq_Bernoulli}
\hat{g}_t \sim \text{Ber}(\text{Sig}(r_{t_2}-r_{t_1}))
\end{equation}
where $\text{Ber()}$ represent Bernoulli distribution and $\text{Sig()}$ represents the sigmoid function. 

\par
According to Eq.~(\ref{eq_Bernoulli}), the sigmoid of the reward difference will be greater than $0.5$ when $r_{t_2}-r_{t_1}>0$, which results in a higher probability of returning a comparison outcome $\hat{g}_t=1$. To train PTR in an end-to-end fashion, the training loss function of PTR $\mathcal{L}_{ptr}$ consists of the reconstruction loss of VAE and cross-entropy loss of the pairwise image comparison outcome as follows:
\begin{equation} 
\begin{split} \label{eq_Loss_Michael}
\mathcal{L}_{ptr} &= \mathcal{L}_{V}(t_1) + \mathcal{L}_{V}(t_2) + \mathcal{L}_{rank}^{t} \\
\mathcal{L}_{V}(\tau) &= -\underbrace{\mathbb{E}_{z_{\tau}\sim q(z_{\tau}|x)} \left[\log p(x|z_{\tau})\right]}_{\text{reconstruction}} + \underbrace{\mathbb{KL}(q(z_{\tau}|x)||p(z_{\tau}))}_{\text{regularization}} \\
\mathcal{L}_{rank}^{t} &= -\frac{1}{N}\sum_{i=1}^N \left[g_t^i \log (\hat{g}_t^i)) + (1-g_t^i) \log (1-\hat{g}_t^i))\right]
\end{split}
\end{equation}
where $\mathcal{L}_{V}(t_1)$ and $\mathcal{L}_{V}(t_2)$ are the CVAE loss functions~\citep{sohn2015learning} for the images recorded at $t_1$ and $t_2$, respectively. $\mathcal{L}_{V}(\tau)$ ($\tau = t_1~\text{or}~t_2$) consists of reconstruction and regularization terms. The reconstruction error is represented by the expected negative log-likelihood of the datapoint, and the regularization error is computed by Kullback-Leibler divergence between the encoder’s distribution $q(z|x)$ and prior distribution $p(z)$.  \final{$\mathcal{L}_{rank}^{t}$ is the probabilistic ranking loss used to enforce that the images presented in the later phase of the demonstrations have higher rewards than those shown in the early phase.} $\hat{g}_t^i$ is the computed comparison output logit using the generative model as Eq.~(\ref{eq_Bernoulli}) for pairwise observations, and $g_t^i$ is the comparison output ground truth of $i$-th images pair.

\par
Since the image comparisons can only be generated internally among each demonstration, PTR will assign the highest reward to each of the demonstrations ending frames. Therefore, PTR is theoretically only able to search for the ``point" object, where the target US plane can only be visualized at a unique pose. Compared with the concept of a ``point" object, a ``line" object means the standard planes can be achieved in multiple points. The standard US plane for a ``line" object is also often needed in clinical practice, such as the longitudinal view of vascular structures for estimating blood flow for PAD diagnosis. In addition, without considering the global information, the generated temporal image comparisons are biased, which will hinder the accurate understanding of the expert's intention.

\subsection{Expert Demonstrations Evaluation and Cleaning}
\par
\revision{Since biased data would mislead the correct understanding of the expert's intention, we propose a probabilistic approach to evaluate the quality of recorded demonstrations and filter out the undesired ones from training data. A confidence value between $[0,1]$ will be assigned to each demonstration based on the positional information of the final frames in all demonstrations. }

\par
Considering the task of visualizing standard US planes, a demonstration ending at a good viewpoint, where the standard planes could be displayed, is considered as a good demonstration. Since the resulting B-mode imaging depends on the end-effector pose, the end-effector poses of the last frames in all demonstrations were employed as the surrogate of the corresponding US images. Then, the last end-effector poses $\textbf{P}_p\in R^6$ were modeled using a joint Gaussian distribution with $N_{dof}$ (the number of DoFs) variables, and the parameters of the distribution were determined using a maximum likelihood method. 

\begin{equation}\label{eq_joint_gaussian}
\textbf{P}_p \sim \mathcal{N}(\pmb{\mu}_g, \pmb{\Sigma}_g)
\end{equation}
where $\pmb{\mu}_g = \mathbb{E}(\textbf{P}_p)$ and $\pmb{\Sigma}_g$ are the mean vector and covariance matrix of $\textbf{P}_p$ from all demonstrations, respectively. $\Sigma_{i,j} = \mathbb{E}[(P_p^i - \mu_g^i)(P_p^j - \mu_g^j)]$, $1\leq i,j \leq N_{dof}$. To balance the numerical difference in different DoFs, $\textbf{P}_p$ has been normalized into $[0, 1]$ in each DoF.

\par
Regarding the multivariate normal distribution, the confidence interval for a certain significance level $p_{ci}$ results in a region, in which $\textbf{P}_p$ satisfies the following condition.

\begin{equation}\label{eq_confidence_interval}
(\textbf{P}_p-\pmb{\mu}_g)^T\pmb{\Sigma}_g^{-1}(\textbf{P}_p-\pmb{\mu}_g)\leq \chi_{N_{dof}}^2(p_{ci})
\end{equation}
where $\chi_{N_{dof}}^2$ is the percent-point function for probability $p_{ci}$ of the chi-squared distribution with $N_{dof}$ degrees of freedom. If the last pose of a sampled demonstration is not inside the confidence region with a given significance level (i.e., $5\%$), this demonstration is considered different from others and should be removed from the data set.




\subsection{Data Augmentation}
\par
Data augmentation is one of the most commonly used regularization methods. It prevents the model from overfitting and increases generalizability. During the training phase, the input data will be randomly changed by a set of transformation functions $T_n^{(p_i,\alpha_i)}$, where $p$ and $\alpha$ are probability and magnitude, respectively. 

\begin{equation}\label{eq_img_transformaitons}
\hat{x}_i = T_n^{(p_n,\alpha_n)}(T_{n-1}^{(p_{n-1},\alpha_{n-1})}(\cdots T_1^{(p_1,\alpha_1)}(x_i)))
\end{equation}
where $n=7$ is the number of used transformations. The detailed transformations are discussed as follows: 

\subsubsection{\final{Image Style Transformations}}
\par
Domain shift in US images is often observed in daily US practice. US images acquired from the same patients using different US machines or acquisition settings will lead to different imaging styles. Due to inter-patient variations, US images of the same anatomy also appear visually different. One data-efficient solution for this is to do style augmentations to the training data. The transformations of \textit{blurriness}, \textit{sharpness}, \textit{Gaussian noise}, \textit{brightness}, and \textit{contrast} are involved in this work. For blurriness, Gaussian filtering is used, where the magnitude $\alpha$ is defined as the standard deviation of the Gaussian filter, ranging between $[0.25,1.5]$. The augmentation in sharpness is done by implementing the unsharp masking technique, where the magnitude $\alpha$ ranges between $[10,30]$. Gaussian noise with zero mean and a standard deviation between $[1,10]$ is used to generate noised images. The brightness of the image is transferred in the range of $[-25,25]$, while the contrast of the image is manipulated using gamma correction with a gamma value ranging between $[0.5,3]$. All transformation functions have the same probability $p$ of $10\%$.

\subsubsection{\final{Image Spatial Transformations}}
\par
Apart from the style transformations, the spatial augmentations of US images should also be considered, in particular for US images. Two types of spatial transformation are involved: \textit{crop} and \textit{flip}. In some scenarios, like vessels, the horizontal position of the target anatomy is linked to the ranking results. Therefore, the crop is only carried out in the vertical direction. The magnitude, in this case, is the scaling factor ranging between $[0.8,0.9]$, while the probability $p$ is $50\%$. Flipping the images horizontally can also increase the variations of limited training images. The probability of flipping is set to $p=10\%$.


\subsection{Global Probabilistic Spatial Ranking (GPSR)} ~\label{sec:III-GPSR}

\par
Regarding PTR, a few limitations have been discussed in Section~\ref{sec:III-PTR}, e.g., biased training comparisons and the inability to be used for ``line" targets. To further overcome these challenges, we consider using spatial ranking instead of temporal ranking. Since the US images, timestamp, and probe pose are paired, the temporal assumption used in Section~\ref{sec:III-PTR} can be adapted as assuming that the US images observed at the position closer to the end pose should have a greater reward without further claims. Since the temporal information is unique and non-repeatable, the time-based image comparisons can only be generated from individual demonstrations, which could result in biased training data. Without global pairwise comparisons, the difference between the last observations of all involved demonstrations will limit the achievement of a precise and unbiased result. In the worst case, inappropriate comparisons will decay the reward inference capability of the PTR approach. In contrast, physical location is repeatable. The use of spatial information enables the generation of unbiased comparisons from all experts' demonstrations. Since all demonstrations are used, the number of training data (pairwise comparisons) generated from a few demonstrations will be soaring. Compared to the temporal differences, the differences between the probe poses of the two images are more intuitive and suitable to represent the difference between the current US image and the target images. Ranking based on spatial information is especially valuable for sub-optimal expert demonstrations, where the target US planes could be witnessed multiple times during a single acquisition.

\par
To display the US plane of the target anatomy, a US probe needs to be controlled in four DoFs for most applications (``point object"), e.g., the task of visualizing and horizontally centering the lamb kidney and chicken heart (Fig.~\ref{Fig_standard_plane}). Two translational movements in the plane orthogonal to the probe centerline and two rotational movements along the probe long axis and the probe centerline, respectively. The remaining two DoFs: the translation along the probe centerline and the rotation around the probe short axis, only change the visualized part of the same plane. To obtain high-quality US images, the probe is required to make firm contact with patients' skin. To ensure patient safety and avoid significant variation of image deformation during scans, a compliant controller~\citep {jiang2021autonomous, hennersperger2016towards} is employed to maintain a constant contact force in the probe centerline direction. Besides, for some anatomies, i.e., tubular structures, the translational position along the centerline of tubular structures is flexible to display a longitudinal vascular view (``line object").

\par
For the proposed GPSR, the spatial cue of each frame is used as the supervisory signal to generate image comparisons among all demonstrations. Considering the ``line" object, target US planes could be achieved at different locations, and the relative positional and rotational differences are computed internally for each demonstration. The use of relative spatial difference $\textbf{d}_i^k$ ($i$-th frame in $k$-th demonstration) instead of absolute difference also allows for using mixed demonstrations obtained at different time and even different patients.



\begin{equation}\label{eq_pose_difference}
\textbf{d}_i^k = \left[|\textbf{P}_i^k(1)-\textbf{P}_e^k(1)|, ..., |\textbf{P}_i^k(6)-\textbf{P}_e^k(6)|\right]
\end{equation}
where $\textbf{P}_i^k\in R^6$ and $\textbf{P}_e^k\in R^6$ are the corresponding probe poses of $i$-th frame and ending frame of $k$-th demonstration. 

\par 
To generate global comparisons between frames from different demonstrations, the computed $\textbf{d}_i^k$ of all images are used to compute the global maximum pose difference vector in all six DoFs $\textbf{d}_{max}^g = \left[\max{(d_i^k(1))},..., \max{(d_i^k(6))} \right]$. To balance the numerical difference between translational and rational movements, $\textbf{d}_i^k$ is further normalized to $[0,1]$ in each DoF, respectively, as follows:

\begin{equation}\label{eq_normalization}
\mathfrak{D}_i^k = \left[\frac{\textbf{d}_i^k(1)}{\textbf{d}_{max}^g(1)}, ..., \frac{\textbf{d}_i^k(6)}{\textbf{d}_{max}^g(6)}\right]
\end{equation}

\par
The generalized global distance $D_i^k$ is computed as Eq.~(\ref{eq_generalized_distance}).

\begin{equation}\label{eq_generalized_distance}
\begin{split}
D_i^k &= \sqrt{\sum_{j=1}^{6}\left[k_j\left(\mathfrak{D}_i^k(j)\right)^2\right]} \\
\text{where}~k_j &= \left\{\begin{split}
    \frac{\left(\textbf{d}_{max}^g(j)\right)^2}{\sum_{m=1}^{3}\left(\textbf{d}_{max}^g(m)\right)^2}~k_t~~~~j=1,2,3\\
    \frac{\left(\textbf{d}_{max}^g(j)\right)^2}{\sum_{m=4}^{6}\left(\textbf{d}_{max}^g(m)\right)^2}~k_r~~~~j=4,5,6\\
\end{split}
\right.
\end{split}
\end{equation}
where $k_t$ and $k_r$ are the weights for translational and rotational movements, respectively, $k_t+k_r = 1$. $k_t=k_r=0.5$ in this study. $k_j$ is the weight for the movement in $j$-th DoF, which is computed based on the $\textbf{d}_{max}^g$ determined by the demonstrations. The DoF with larger positional/rotational variations will result in a larger weight. 

\begin{figure}[ht!]
\centering
\includegraphics[width=0.45\textwidth]{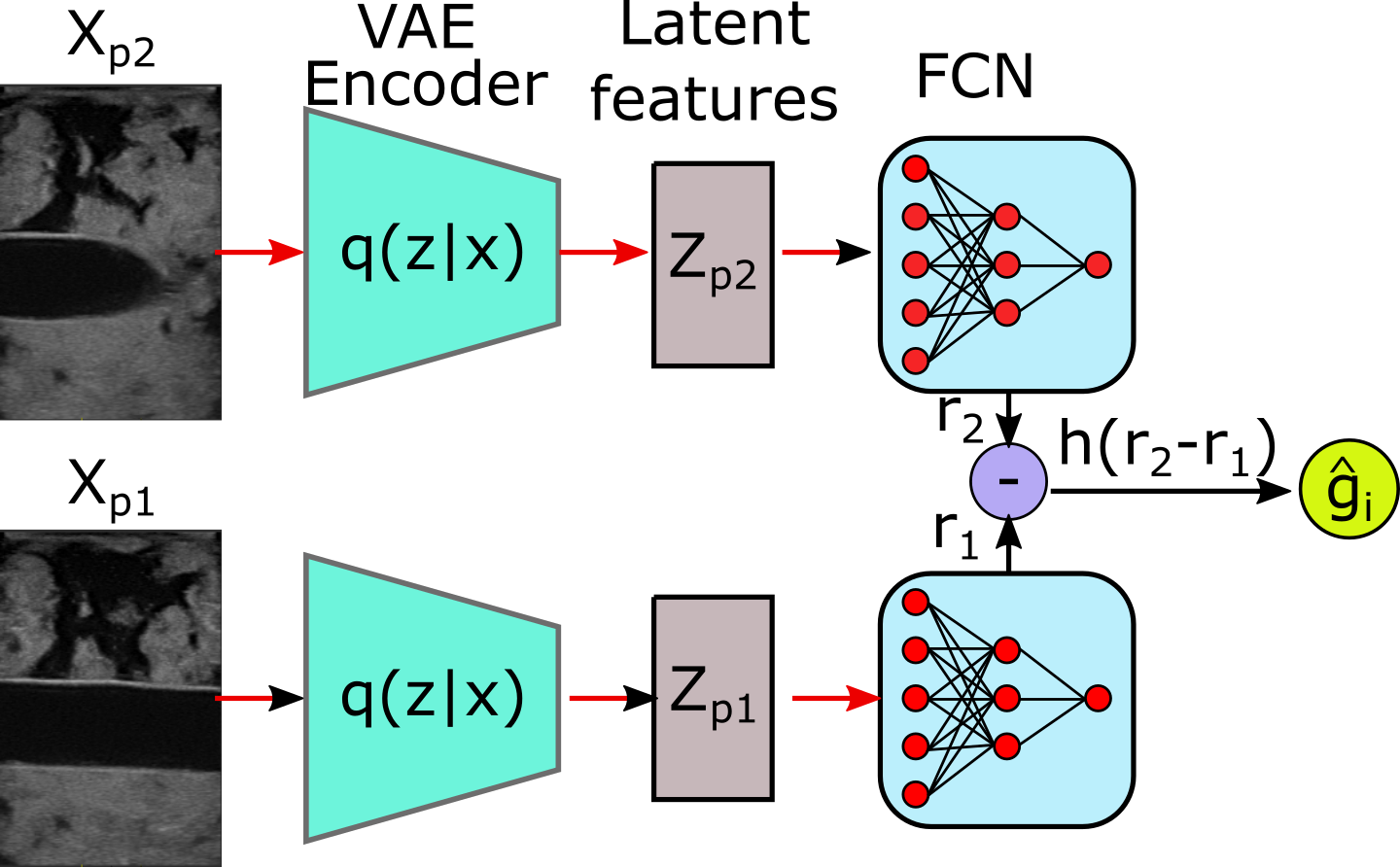}
\caption{The architecture of the GPSR reward network. The paired images are encoded by a pre-trained VAE. Then, the image features ($Z_{p}$) in latent space are fed to an FCN $r_{\psi}(Z)$ to predict the reward for individual images. Based on the predicted rewards for paired images, the comparison outcome can be estimated using $h(*)$.  
}
\label{Fig_RewardNetStructure_GPSR}
\end{figure}

\par
Based on the computed generalized distance $D_i^k$ for individual images, the ground truth of global comparison outcomes is automatically determined. For pairwise images ($X_{p1}$ and $X_{p2}$) randomly sampled from demonstrations, the comparison outcome $g=1$ if $D_i^k (p_1)\geq D_i^k(p_2)$, and $g=0$ if $D_i^k (p_1)< D_i^k(p_2)$, where $p_1$ and $p_2$ are two random probe poses that appeared in the given demonstrations. 

\par
The structure of the GPSR is detailed in Fig.~\ref{Fig_RewardNetStructure_GPSR}. The VAE encoder $q(z|x)$ is pre-trained on a large data set to make GPSR adaptable to different image styles. Then, the extracted features in latent space are directly fed to an FCN reward prediction network $r_{\psi}(Z)$. To optimize the parameters of the FCN, the binary cross entropy loss over the comparison outcome is used as reward loss function $\mathcal{L}_{rank}^{s}$.


\begin{equation}\label{eq_loss_rank}
\mathcal{L}_{rank}^{s} = -\frac{1}{N}\sum_{i=1}^N \left[g^i_{s} \log (\hat{g}^i_{s}) + (1-g^i_{s}) \log (1-\hat{g}^i_{s})\right]
\end{equation}
where $\hat{g}_{s}$ is predicted pairwise comparison outputs for images obtained at different locations $p_1$ and $p_2$, respectively. 

\begin{equation}\label{eq_Bernoulli_gpsr}
\hat{g}_{s} \sim \text{Ber}\{\text{Sig}\left[k_{re}(r_{p2}-r_{p1})\right]\}=h(r_{p2}-r_{p1})
\end{equation}
where $k_{re}$ is a coefficient used to restrain false high reward. Here, $k_{re}$ is empirically determined as $10$ based on the experimental performance.

\subsection{Mutual Information-based Feature Disentanglement} ~\label{sec:III-MI}
\par
In this work, feature disentanglement requires task-related features $Z^r$ and domain features $Z^d$ to have their specific information, while they do not incorporate the information of others. Such explicit feature disentanglement will improve the generalization capability of unseen images by removing the potential disturbs caused by unseen domain information from the task-related features. To this end, we need to quantitatively estimate the dependence between $Z^r$ and $Z^d$. Mutual Information (MI) which measures the amount of information obtained from one random variable by observing the other random variable, is exactly the quantity we need. The definition of MI is defined as follows:

\begin{equation}\label{eq_MI}
I(Z^r;Z^d) = \int_{\mathbb{R}\times \mathbb{D}} {\log{\frac{d\mathbb{P}_{rd}}{d\mathbb{P}_r \otimes \mathbb{P}_d}}d\mathbb{P}_{rd}}
\end{equation}
where $\mathbb{D}$ and $\mathbb{R}$ are two distributions of variables $Z^d$ and $Z^r$, respectively. $\mathbb{P}_{rd}$ is the joint probability distribution of $(Z^r, Z^d)$, $\mathbb{P}_d=\int_{\mathbb{R}}\mathbb{P}_{rd}$ and $\mathbb{P}_r=\int_{\mathbb{D}}\mathbb{P}_{rd}$ are the marginals. $\mathbb{P}_r \otimes \mathbb{P}_d$ is the product of the marginal distributions. However, the fast computation of MI is limited to the discrete samples~\citep{paninski2003estimation}. Thereby, MINE~\citep{belghazi2018mutual} is used to approximate the lower bound of MI based on limited samples by a neural network $T_{\theta}$, which is used to substitute a random distribution. To avoid the computation of the integrals practically, Monte-Carlo integration is applied to compute the lower bound of the MI $I(\widehat{Z^r; Z^d})$.

\begin{equation}\label{eq_MI_approximate}
I(\widehat{Z^r; Z^d}) = \frac{1}{n}\sum_{i=1}^{n}T_{\theta}(Z^r, Z^d) - \log{\frac{1}{n}\sum_{i=1}^n{e^{T_{\theta}(Z^r, \hat{Z}^d)}}}
\end{equation}
where $(Z^r, \hat{Z}^d)$ are samples sampled from the product of marginal distributions $\mathbb{P}_r \otimes \mathbb{P}_d$, while $(Z^r, Z^d)$ are sampled from joint distributions $\mathbb{P}_{rd}$.

\begin{figure}[ht!]
\centering
\includegraphics[width=0.45\textwidth]{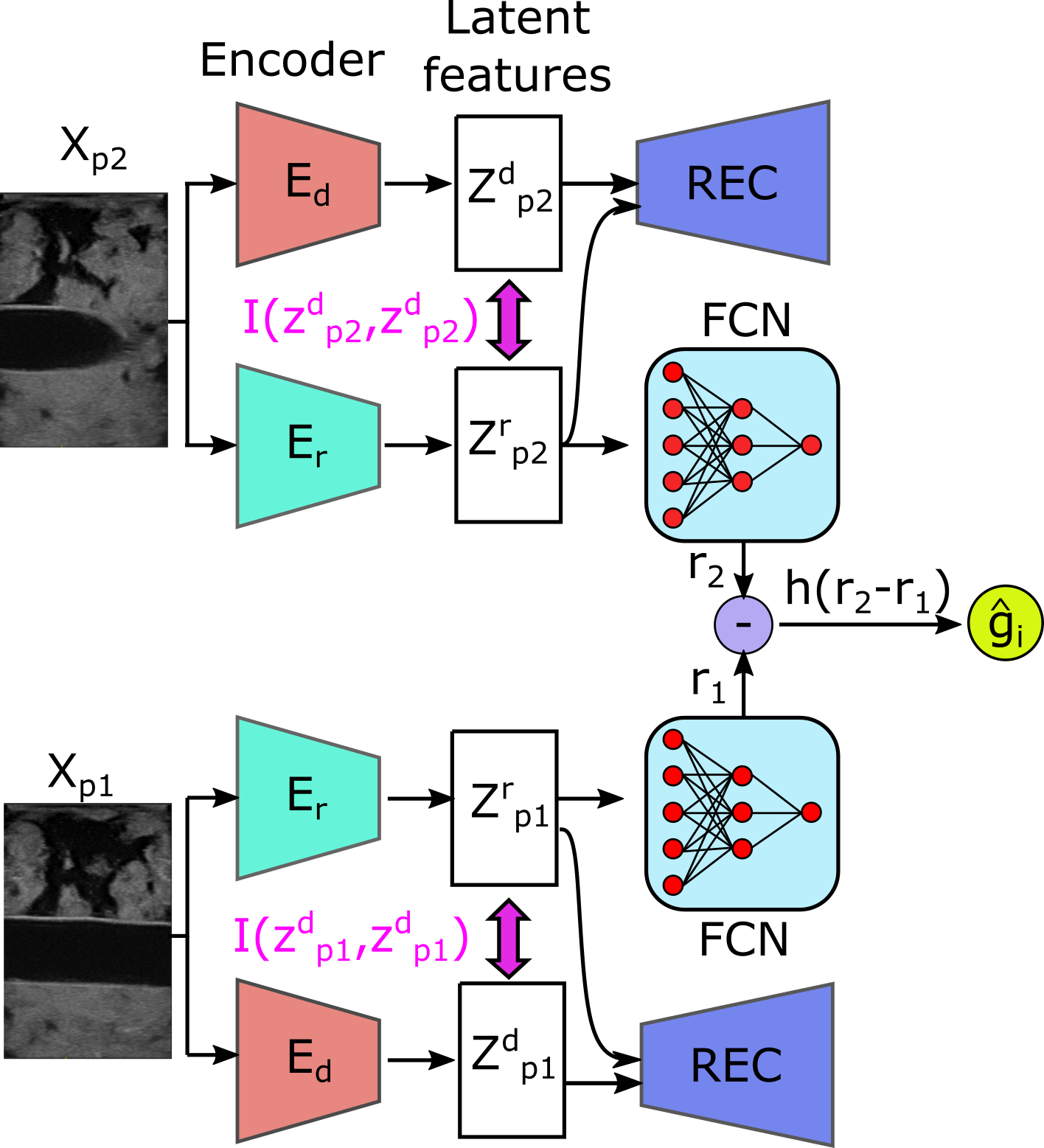}
\caption{The architecture of the proposed mutual information-based disentangled reward network (MI-GPSR). The inputs are image pairs randomly sampled from demonstrations. For each image, two independent encoders ($E_d$ and $E_r$) are used to extract latent features. To explicitly enforce the disentanglement between task-related features $Z^r$ and domain features $Z^d$, their MI term is minimized during the training process. To ensure the latent features can represent the inputs, $Z^r$ and $Z^d$ are concatenated to compute the reconstructed images. To train the reward network FCN, the pairwise $Z^r_{p1}$ and $Z^r_{p2}$ are fed as input to estimate rewards $r_1$ and $r_2$. The parameters of FCN are optimized based on the automatically generated comparison outputs $g$. The same background color of encoders, decoders, and FCN indicates that they have shared weights. 
}
\label{Fig_MI_RewardNetStructure}
\end{figure}

\subsection{Training of the Proposed MI-GPSR Framework}

\par
Based on the MINE, the architecture of the novel MI-based disentangled reward network (MI-GPSR) is depicted in Fig.~\ref{Fig_MI_RewardNetStructure}. To train the MI-GPSR, three different loss functions are employed: (1) image reconstruction loss $\mathcal{L}_{rec}$, (2) MI neural estimator loss $\mathcal{L}_{MI}$, and (3) spatial information-based pairwise image ranking loss $\mathcal{L}_{rank}^{s}$.

\subsubsection{Image Reconstruction Loss}
The MI-GPSR is implemented in the architecture of Encoder-Decoder where two encoders independently extract latent representations from the original input $X_p$. Two encoders $E_r$ and $E_d$ are used to respectively extract task-related features $Z^r$ and domain features $Z^d$, where $Z^r =E_r(X_p, \theta_{Er})$ and $Z^d =E_d(X_p, \theta_{Ed})$. The decoder $REC$ is employed to guarantee the combination of features $Z^r$ and $Z^d$ can recover the original input images, where $\hat{X}_p =REC(Z^r, Z^d, \theta_{REC})$. Here, $\theta_{Er}$, $\theta_{Ed}$ and $\theta_{REC}$ are the parameters of $E_r$, $E_d$ and $REC$, respectively. The mean squared error function is employed as the image reconstruction loss $\mathcal{L}_{rec}$ as follows:

\begin{equation}\label{eq_loss_reconstruction}
\mathcal{L}_{rec} = \frac{1}{mn}\sum_{i=0}^{m-1}\sum_{j=0}^{n-1}\left[X_p(i,j)-\hat{X}_p(i,j) \right]^2
\end{equation}
where $i,j$ is the image pixel location, $m$ and $n$ are the image width and height, respectively.

\subsubsection{MI Neural Estimator Loss}
\par
To enforce the explicit feature disentanglement in latent space, MI neural estimator $T_{\theta}$ (see Sec.~\ref{sec:III-MI}) is employed. To optimize the network parameter $\theta_{MI}$ of $T_{\theta}$, we need to maximize the lower bound of MI $I(\widehat{Z^r; Z^d})$ depicted in Eq.(~\ref{eq_MI_approximate}) to achieve a looser estimation of MI. Therefore, the MI neural estimator loss function $\mathcal{L}_{MI}$ can be defined as follows:

\begin{equation}\label{eq_loss_MI}
\mathcal{L}_{MI} = -I(\widehat{Z^r; Z^d})
\end{equation}

\subsubsection{Spatial Information-based Pairwise Image Ranking Loss}
\par
In order to make the network understand the underlying expert ``intention", this work proposes a spatial ranking approach (see Sec.~\ref{sec:III-GPSR}) to train the MI-GPSR in a self-supervised mode. The proposed MI-GPSR also eliminates the requirement of burdensome manual annotation. The expert ``intention" is inferred as reward $r\in[0,1]$, where a high value will be assigned when the image is close to the target standard US planes. To estimate the reward value for individual images, a FCN $FCN_{\theta}$ is used to compute the reward of random input $X_p$ as $r=FCN_{\theta}(E_r(X_p))$. To train the MI-GPSR in a self-supervised mode, pairwise image comparisons are used. 
The spatial information-based pairwise image ranking loss $\mathcal{L}_{rank}^{s}$ is the same as Eq.~(\ref{eq_loss_rank}).




\par
The MI-GPSR is trained in an alternating fashion, the estimated MI $I(Z^r, Z^d)$ is used as the supervision to update both $E_r$ and $E_d$. $\mathcal{L}_{rec}$ is used to update $E_r$, $E_d$ and $REC$. Similarly, $\mathcal{L}_{rank}^{s}$ is used to update $E_r$ and $FCN$ for reward inference. In this work, the batch size is $8$ and the size of latent feature $Z^r$ and $Z^d$ are $32$. The detailed process of MI-GPSR is depicted in Algorithm~\ref{algorithm_network_update}. 
\final{
As the training progresses, $E_r$ is likely to focus on task-related features. Otherwise, $\mathcal{L}_{rank}^{s}$ will increase because of inconsistent predictions of reward for paired image comparisons. In addition, by minimizing the MI computed between $Z^d$ and $Z^r$ in latent space, $E_d$ will be prone to only contain the domain features, which are complementary to task-specific features.  
}

\begin{algorithm}[htb] 
\caption{Training process of MI-GPSR reward network}~\label{algorithm_network_update}
\begin{algorithmic}[1]
\State $\theta \gets$ initialize MI-GPSR parameters
\Repeat
    \State ($X_{p1}$, $X_{p2}$, $g$) $\gets$ random pairwise comparisons batch 
    \State $Z^r_{p1}\gets E_r(X_{p1})$; $Z^d_{p1}\gets E_d(X_{p1})$; $Z^r_{p2}\gets E_r(X_{p2})$; $Z^d_{p2}\gets E_d(X_{p2})$; $\hat{X}_{p1}\gets REC(Z^r_{p1}, Z^d_{p1})$; $\hat{X}_{p2}\gets REC(Z^r_{p2}, Z^d_{p2})$
    \State $\left[\mathcal{L}_{MI}(X_{p1}), \mathcal{L}_{MI}(X_{p1})\right]\gets$ Eq.~(\ref{eq_loss_MI})
    \State //update MINE based on gradients
    \State $\theta_{MI}\gets \bigtriangledown_{\theta_{MI}}\left(\mathcal{L}_{MI}(X_{p1}) + \mathcal{L}_{MI}(X_{p2}) \right)$
    \State $\left[I(\widehat{Z^r_{P1}; Z^d_{p1}}),I(\widehat{Z^r_{P2}; Z^d_{p2}})\right] \gets$ Eq.~(\ref{eq_MI_approximate}) 
    \State $\left[\mathcal{L}_{rec}(X_{p1}), \mathcal{L}_{rec}(X_{p2})\right]\gets$ Eq.~(\ref{eq_loss_reconstruction})
    \State $\mathcal{L}_{rank}^{s}(X_{p1}, X_{p2})\gets$ Eq.~(\ref{eq_loss_rank})
    \State //update MI-GPSR parameters based on gradients
    \State $(\theta_{E_d}, \theta_{E_r}, \theta_{REC})\gets \bigtriangledown \left(\mathcal{L}_{rec}(X_{p1}) + \mathcal{L}_{rec}(X_{p2}) \right)$
    \State $(\theta_{E_d}, \theta_{E_r})\gets \bigtriangledown \left(I(\widehat{Z^r_{P1}; Z^d_{p1}}) + I(\widehat{Z^r_{P2}; Z^d_{p2}}) \right)$
    \State $(\theta_{E_r}, \theta_{FCN})\gets \bigtriangledown \mathcal{L}_{rank}^{s}(X_{p1}, X_{p2})$
\Until{end}
\end{algorithmic}
\end{algorithm}

\section{Autonomous Navigation to Standard Planes}

\begin{figure}[ht!]
\centering
\includegraphics[width=0.35\textwidth]{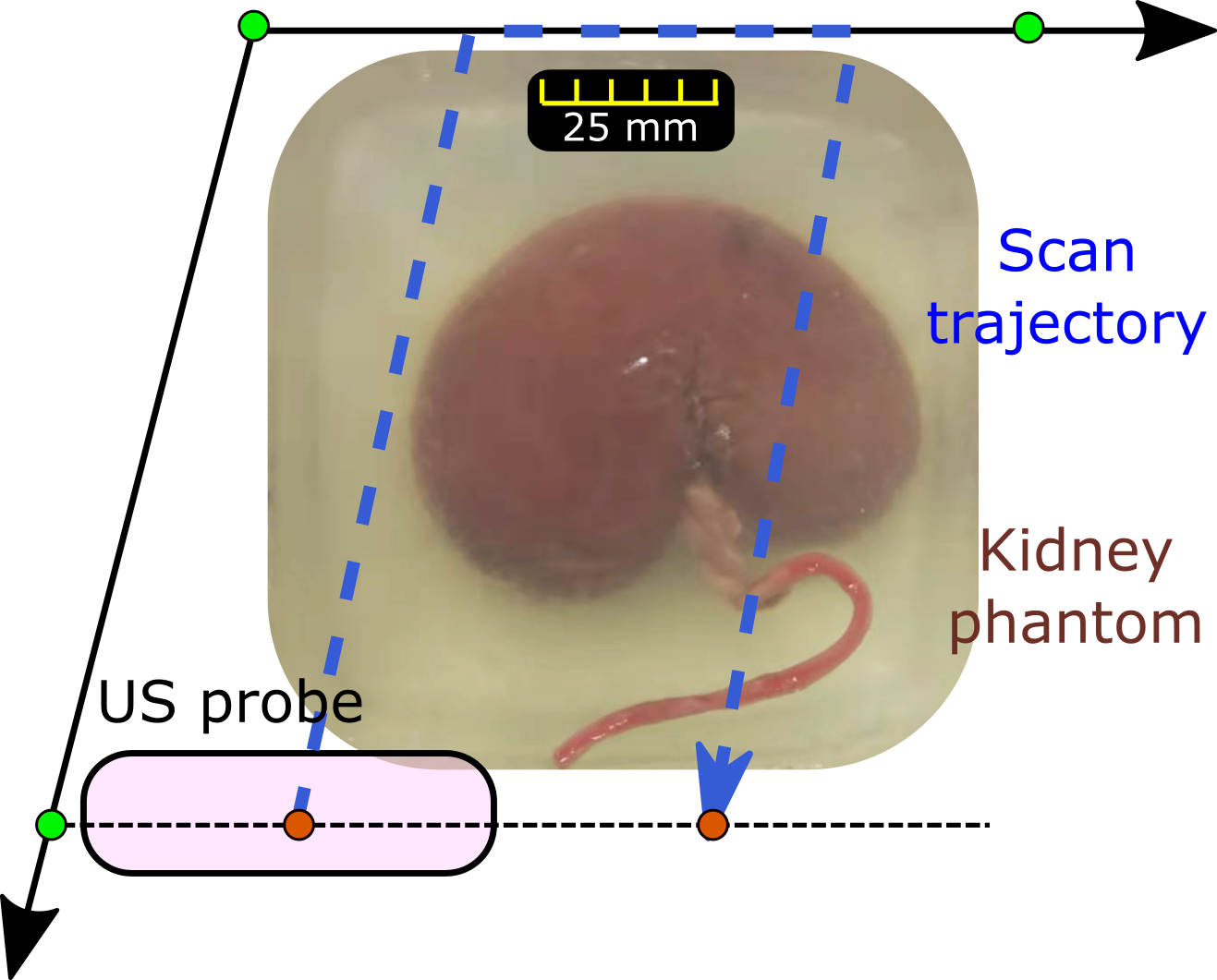}
\caption{Representative scanning trajectory generated for fully covering the anatomy of interest.}
\label{Fig_path_planing}
\end{figure}

\subsection{Generation of Simulated US Images}
\subsubsection{Robotic Scanning and US Volume Reconstruction}
\par
The proposed reward inference network MI-GPSR enables reward estimation of individual images based on the disentangled task-related features $FCN_{\theta}(Z^r)$. Theoretically, it is possible to evolve a policy to navigate a US probe towards target standard planes using RL methods or Gaussian processes. Nevertheless, to find an optimal policy in real scenarios, rich interaction between RUSS and patients are needed to balance both exploration and exploitation. Due to safety concerns and time efficiency, these approaches are not suitable for medical US examination tasks. 

\par
To tackle the practical limitations, a new framework is proposed in this work. Instead of real-time interaction with patients, a single sweep over the anatomy of interest is carried out first. The region of interest is manually defined by operators by selecting three points ($\textbf{P}_i^m$) on the surface of the target in a clockwise or anti-clockwise direction. The target is covered by a parallelogram defined by $\textbf{P}_i^m$. To obtain 3D volumetric target, the multi-line trajectory planning strategy~\citep{huang2018fully} is employed (see Fig.~\ref{Fig_path_planing}). 
To limit the total length of the trajectory, a probe is moved in the direction parallel to the longer edge of the manually defined parallelogram, i.e., $\overrightarrow{p_2^m p_1^m}$ when $|p_2^m p_1^m| \geq |p_2^m p_3^m|$. The key points $\textbf{P}_k$ required to define the multi-line trajectory are calculated as follows:

\begin{equation}\label{eq_path_key_points}
\textbf{P}_k = \left\{
\begin{split}
&\textbf{P}_2^m+\frac{1}{2}\overrightarrow{p_2^m p_i^m}~~~~w_p\geq \text{min}\{|p_2^m p_1^m|, |p_2^m p_3^m|\}\\
&\textbf{P}_2^m+\left[\left(\frac{1}{2}+j \right)w_p -j\epsilon_0\right]\frac{\overrightarrow{p_2^m p_i^m}}{|\overrightarrow{p_2^m p_i^m}|}~~~\text{Others}\\
\end{split}
\right.
\end{equation}
where $\overrightarrow{p_2^m p_i^m}$ is the axis with a shorter length, $i=\{1, 3\}$, $j = 1, 2, ..., \lceil \frac{\text{min}\{|p_2^m p_1^m|, |p_2^m p_3^m|\}}{w_p} \rceil$, $\epsilon_0$ is a small coefficient used to guarantee the overlap between the sweep along two neighbouring lines.

\par
Based on the planned trajectory, the tracked B-mode images can be stacked spatially in 3D space based on the known robotic kinematic model. To generate 3D volume, a linear interpolation was employed using PLUS~\citep{lasso2014plus}.

\begin{figure}[ht!]
\centering
\includegraphics[width=0.45\textwidth]{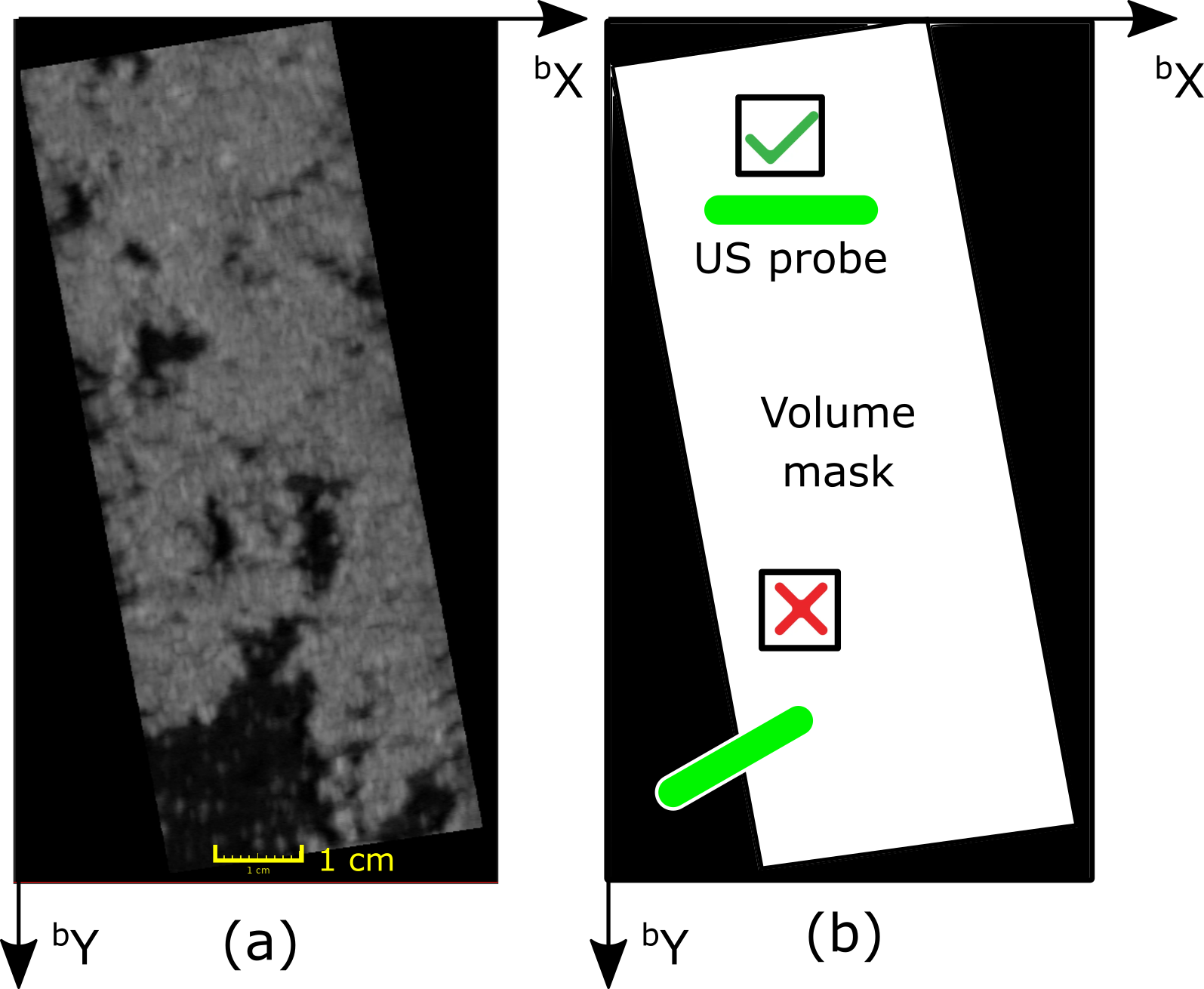}
\caption{\revision{Illustration of generating simulated images from a recorded volume. (a) the projected view ($X\mbox{-}Y$ plane of robotic base frame) of a representative US volume. (b) the masked version of the projected view.}}
\label{Fig_projected_mask}
\end{figure}

\subsubsection{Simulated US Images from Real Volume}
\par
\final{To ensure patient safety, only one US sweep over the target is carried out. Since the US standard plane can only be achieved by correctly positioning and orienting the probe, the standard US plane may not be contained in the recorded US sweep. Compared to taking multiple sweeps on patients, simulating 2D US images with varying acquisition parameters from the compounding volume can reduce the dynamic contact with patients and time to get images with different acquisition parameters. Thereby it can minimize the risk for patients and improve the efficiency of the US examination procedure.}
To obtain 2D images from the compounding result, a virtual rectangle ($w_p\times d_p$) is employed to mimic US imaging plane, where $w_p$ and $d_p$ represent probe footprint width and the acquisition depth, respectively. Afterward, the US volume is projected into $X\mbox{-}Y$ plane of the robotic base coordinate system \{b\} \revision{[Fig.~\ref{Fig_projected_mask}~(a)]}. The detailed coordinate systems of the setup can be found in our previous publications~\citep{jiang2021autonomous}. The probe pose in the projected 2D view can be defined as ($^{b}x$, $^{b}y$, and $^{b}R_z$). In case the desired plane is only able to be seen when the probe is tilted, additional rotation around probe long axis ($^{p}X$) $^{p}R_x$ is defined. Since the effect of the rotation around the probe short axis ($^{p}Y$) could be partly replaced with the movement in \revision{$^{p}X$ direction}, this rotation is kept unchanged.

\par
By tuning the probe pose ($^{b}x$, $^{b}y$, $^{b}R_z$, $^{p}R_x$) in a virtual environment, a large number of simulated images could be generated in a shorter time (around $10~ms$ per image) compared to physical acquisitions. The detailed parameters involved here are: 1) $^{b}x = [x_{min}, x_{s}, x_{max}]$, 2) $^{b}y = [y_{min}, y_{s}, y_{max}]$, 3) $^{b}R_z = [0, R_z^{s}, 180^{\circ}]$, and 4) $^{p}R_x = [-30^{\circ}, R_x^{s}, 30^{\circ}]$. The left, middle and right values represent lower boundary, the increment of each step, and the upper boundary, respectively. In this study, $x_{s} = y_{s}=3~mm$ and $R_z^{s}=R_x^{s} = 5^{\circ}$.

\par
To filter out the simulated images without meaningful context (part/full of the simulated image is out of the recorded real volume), the recorded volume mask $M_v$ (2D) was generated by projecting the 3D compounding volume onto $X\mbox{-}Y$ plane of frame \{b\} \revision{[Fig.~\ref{Fig_projected_mask}~(b)]}. The white area ($I=1$) represents the projected volume with context inside, while the black area ($I=0$) is the area padded during 3D compounding. Besides, another mask representing a probe $M_p$ with various acquisition parameters ($^{b}x$, $^{b}y$, and $^{b}R_z$) can be seen as a binary mask with a line on it (see green line in Fig.~\ref{Fig_projected_mask}). The line mask length is defined as the probe width $w_p$. Then, an element-wise multiplication is carried out ($M_v\odot M_p$) to determine whether the current probe line is fully located inside the volume mask $M_v$. Only when the probe line is fully inside $M_v$ (see Fig.~\ref{Fig_projected_mask}), the resulting simulated image will be kept.

\subsection{Alignment of US Probe}
\subsubsection{Coarse Positioning of US Probe}
\par
\begin{figure}[ht!]
\centering
\includegraphics[width=0.48\textwidth]{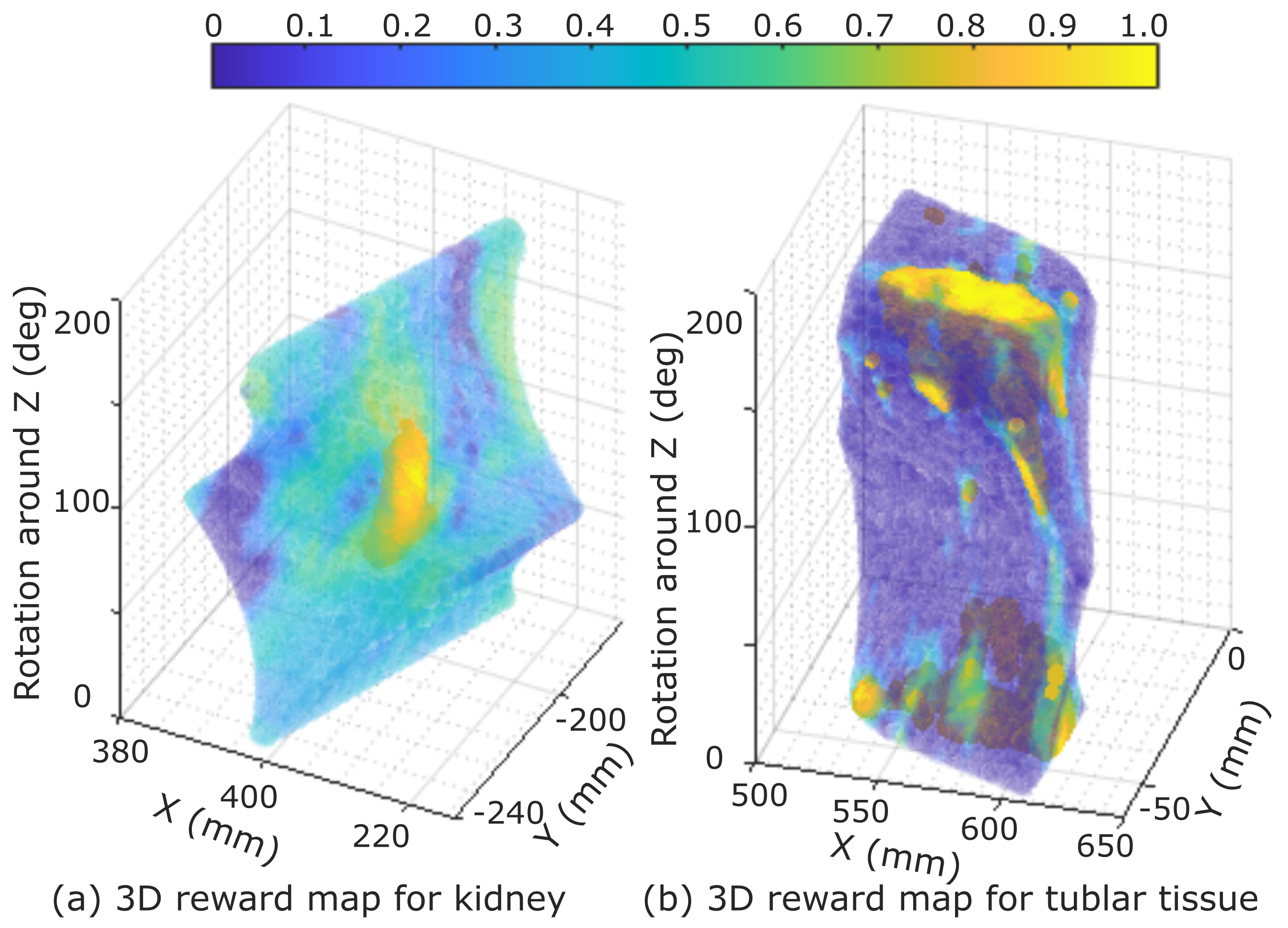}
\caption{Illustration of two inferred reward volumes of (a) ex-vivo kidney phantom and (b) tubular phantom.}
\label{Fig_reward_3d}
\end{figure}

Due to the explicit feature disentangle process, the well-trained reward inference network MI-GPSR $FCN_{\theta}(Z^r)$ can be directly applied to the unseen simulated images (potential with unrealistic artifacts). Thereby, a corresponding reward volume map can be achieved by exhaustively creating simulated images using varying probe poses in the virtual environment.  
To intuitively visualize the reward distribution, two representative reward volumes of tubular phantom and ex-vivo kidney phantom are depicted in Fig.~\ref{Fig_reward_3d}. The tubular phantom and kidney phantom are used to mimic two common navigation task for a ``point" object and ``line" object, respectively.  
In the 3D reward volumes, blue means low reward while yellow represents high reward. A higher reward means being closer to the desired US plane.
It can be seen from Fig.~\ref{Fig_reward_3d} that the high reward ($>0.85$) is mainly distributed around the area ($mean\pm STD$) $[410\pm0.8~mm, -219\pm1.8~mm, 106.5\pm3.1^{\circ}]$ for kidney phantom and $[579\pm17.1~mm, -38\pm3.3~mm, 170\pm6.4^{\circ}]$ for the tubular phantom. For most applications, i.e., the lamb kidney and chicken heart, the highest rewards are concentrated to a point, while for tubular structure these are mainly distributed along a line. This is caused by the difference in the targets' geometric features; the longitudinal view of the blood vessel allows flexibility in the direction of the vessel centerline. The resulting shape of reward volume in Fig.~\ref{Fig_reward_3d} (a) is ``rhombus"  because only the simulated images obtained when the virtual probe is fully inside the volume mask (the white region in Fig.~\ref{Fig_projected_mask}) will be kept. The virtual probe orientation in Fig.~\ref{Fig_projected_mask} is determined by the rotation angle ($^{b}R_z$). When the virtual probe is rotated to be parallel to the short edge of the white region, the valid positions of the probe is restricted within a small area around the center line along the long edge. In contrast, the probe can be moved in a relatively larger area, when the probe is parallel to the long edge. Thereby, more simulated images can be obtained when $^{b}R_z$ is around $90^{\circ}$, while fewer images are generated when $^{b}R_z$ is around zero or $180^{\circ}$ in Fig.~\ref{Fig_reward_3d} (a). Due to the vessel phantom being positioned differently from the kidney phantom, fewer images are obtained when $^{b}R_z$ is around $90^{\circ}$, while more images are obtained when $^{b}R_z$ is around zero or $180^{\circ}$ in Fig.~\ref{Fig_reward_3d} (b). 


\par
To avoid the effect of the potential inaccurate reward prediction, a 3D moving window is employed to smooth the 3D volumetric rewards. Afterward, the simulated image with the highest average reward value is picked, and its acquisition parameters are considered as the coarse location of the target pose displaying the target US plane.
\final{During the approaching process of the identified coarse location, the robotic manipulation is controlled in the impedance control mode to maintain the contact force in the direction of the probe centerline, which is achieved by setting a stiffness and desired force. More detailed implementation of this controller can be referred to~\citep{hennersperger2016towards, jiang2021autonomous}. In order to accurately position the probe to the recapped location from the recorded US volume in the current setup, a relatively larger stiffness is used for other DoFs of the robot ($1000~N/m$ and $20~Nm/rad$ for the translational and rotational movement, respectively). }
Finally, a safety limitation is set to $25~N$ to restrict the maximum force exerted by the robot. If the contact force is larger than $25~N$, the robot will automatically stop to avoid excessive force.

\subsubsection{Fine Adjustment of US Probe}
\par
To mimic human experts' behavior and bridge the potential gap between simulated US images and real images, a fine-tuning process is performed after coarse positioning of US probe. Considering the size of objects, \final{a searching area ($^{b}x:\pm 10~mm$, $^{b}y:\pm 10~mm$, $^{b}R_z:\pm 10^{\circ}$, and $^{p}R_x: \pm 10^{\circ}$) is defined around the coarse pose determined by simulated images. The step size for the involved translational movement was $5~mm$ and the one for the rotational movement was $5^{\circ}$.} The searching procedure is automatically performed inside the predefined area. During the fine-tuning procedure, the reward for each image was computed in real-time. The US image with the maximum reward was finally considered as the best pose to visualize the target standard plane. In addition, the searching parameters used in the fine-tuning procedure can be changed according to the requirements (i.e., accuracy and/or time) of targets with varying sizes.

\section{Results}



\begin{figure*}[ht!]
\centering
\includegraphics[width=0.9\textwidth]{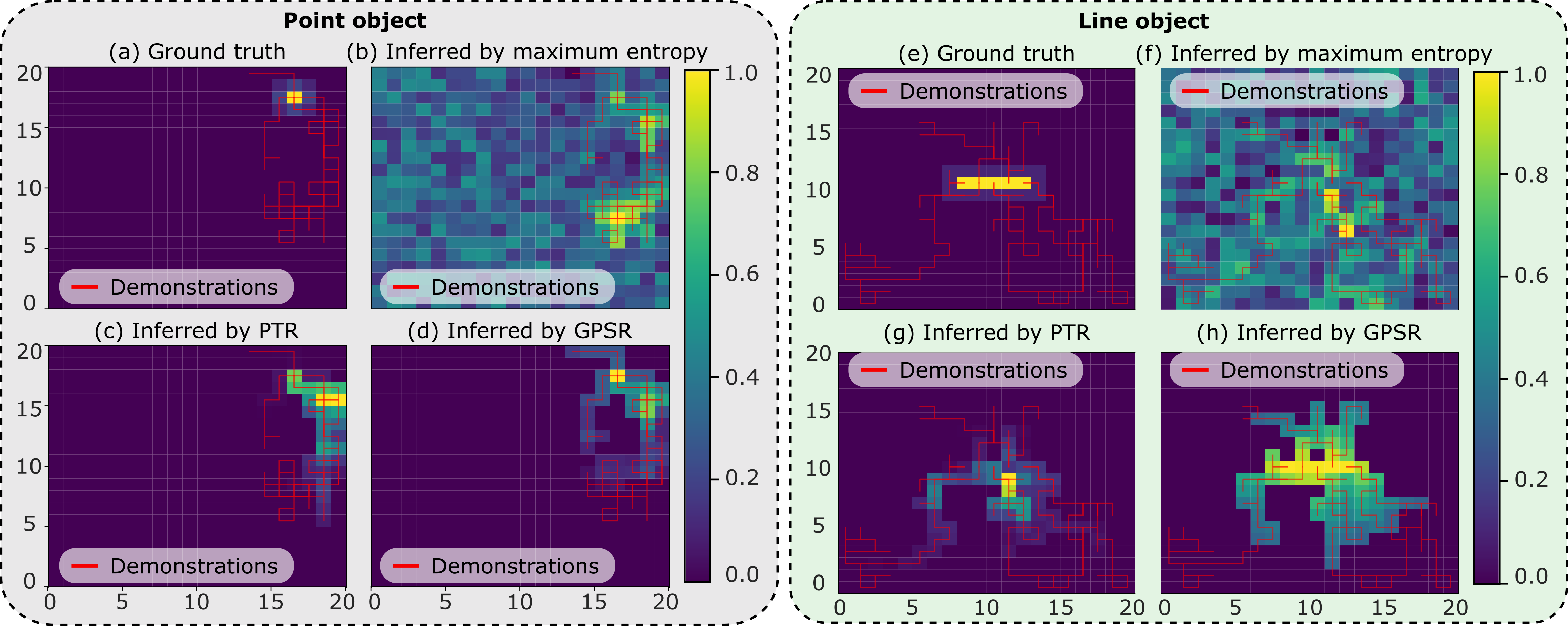}
\caption{Reward maps obtained from sub-optimal demonstrations in grid worlds. Red lines represent the given demonstrations in the grid world, and the color in each grid shows the \revision{inferred} reward of the corresponding position. \revision{The predicted reward value ($[0,1]$ in yellow) reflects the distance between each position and the target grid position/positions.} 
} 
\label{Fig_gridworld_suboptimal}
\end{figure*}

\subsection{Hardware Setup}
The proposed RUSS mainly consists of a robotic arm (KUKA LBR iiwa 7 R800, KUKA Roboter GmbH, Augsburg, Germany) and a US machine (Cephasonics, California, USA). A linear probe (CPLA12875, Cephasonics, California, USA) is rigidly attached to the robotic flange using a 3D printed holder. B-mode images are accessed using a USB interface provided by the manufacturer in $50~fps$.  The robot is controlled using a Robot Operating System interface~\citep{hennersperger2016towards}. The real-time US images and robotic poses were synchronized and further used for 3D compounding in a software platform (ImFusion GmbH, Munich, Germany). The US setting was mainly determined by the default setting file for vascular tissues provided by the manufacturer: brightness: $66~dB$, dynamic range: $88~dB$. The depth, focus and frequency were $5.0~cm$, $3.1~cm$ and $7.6~MHz$ for vascular phantom, while these parameters were changed to $5.5~cm$, $5.0~cm$ and $5.7~MHz$ for the ex-vivo phantoms. 

\par
To validate the performance of the proposed approach for various organs, three types of phantoms (lamb kidney, chicken heart and vessel) were used in this work. The blood vessel phantom was primarily made using gelatin powder, while the ex-vivo organs phantoms were made using candle wax. Candle wax has a good sealing ability, which allows for longer preservation of fresh animal organs. To construct the vascular phantom: gelatin powder ($175~g/L$) was dissolved into water, and the mixed solution was heated to $80$ degree. To mimic human tissue artifacts, paper pulp ($3-5~g/L$) was randomly mixed into the solution. After solidification, a round tube was used to create two holes at different depths of the phantom for mimicking vascular structures. For the ex-vivo organs phantom, hot candle wax liquid was used to cover the organs in a small box. After the liquid was fully solidified, it was taken out and placed in another box. Then, candle wax liquid was poured into the box to submerge the upper surface of the phantom. Like paper pulp used for vascular phantom, ginger powder ($10~g/L$) was mixed with the candle wax for ex-vivo organ phantoms.

\subsection{Performance of GPSR on 2D Grid World}
\par
\revision{To intuitively demonstrate that the proposed method works for both ``point" and ``line" objects, which are the two most representative tasks in clinical practices, a grid world environment ($20\times 20$) is built for qualitative and quantitative analysis. In the grid world, the aim is to move an agent initialized from a random position toward the target position (marked in yellow in Fig.~\ref{Fig_gridworld_suboptimal}).} Since there is no need to disentangle image features in latent space in grid-world environments, the advanced version of GPSR (MI-GPSR) is not tested here.

\par
In order to generate sub-optimal trajectory demonstrations, two q-learning policies were trained separately for the ``point" and ``line" objects in the grid world, in which the reward of the target position was set to one, while the rewards of other positions were zero. The maximum \revision{number of} episodes and steps for each episode were set to $50$ and $100$, respectively.
Based on the sub-optimal trained model, an agent will move toward the target position from a random position. For point object, five demonstrations ending at the target position were generated to mimic standard planes in US applications. In addition, for the line object ($1 \times 5$), ten demonstrations ending at the target line were generated. Since the model was only trained with limited epochs, \final{the generated demonstration was sub-optimal; namely, the trajectory cannot directly move towards the target position in the minimum number of steps.}

\par
Based on the given sub-optimal demonstrations, three reward models were trained using maximum entropy IRL (ME-IRL)~\citep{aghasadeghi2011maximum}, PTR~\citep{burke2023learning} and the proposed GPSR, separately. The ground truth and the reward maps inferred by the three models are shown in Fig.~\ref{Fig_gridworld_suboptimal}. It can be seen from the figure that the reward map inferred by the proposed GPSR is the closest to the ground truth, in which the highest reward clearly shows in the right position in the grid world. The result achieved using the maximum entropy~\citep{aghasadeghi2011maximum} approach failed to correctly recover the reward value from the sub-optimal demonstrations, particularly for the line object. This is mainly because of the inherited limitation of the maximum entropy approach, which tends to assign larger reward values to states observed more often in the demonstration. However, demonstrations of US examination are sub-optimal. The positions observed more often in the demonstrations may not be the desired positions. Regarding PTR approach~\citep{burke2023learning}, the inferred reward map is better than the one obtained from maximum entropy while still worse than the one achieved from the proposed GPSR approach. It can be seen from Fig.~\ref{Fig_gridworld_suboptimal}~(c) that the position with the maximal reward is close to the desired position. However, since the pairwise training data was generated from individual demonstrations, a biased result occurs, especially when the lengths of demonstration are significantly different from each other. For the line object, PTR fails to realize that the object is a line. Only one grid point of the line is successfully assigned with a large reward (see Fig.~\ref{Fig_gridworld_suboptimal}~(g)).

\par
Considering the intrinsic properties of US examination, the target standard planes correspond to a unique or a set of probe poses with a unified character. Thus, the proposed GPSR approach directly generates the paired training data from all demonstrations based on the probe spatial cues rather than temporal cues. Due to the global consideration of all demonstrations, more training data could be generated from the same demonstration using the GPSR approach. The performance of inferred reward in grid world using different approaches also demonstrates the superiority of GPSR over the PTR and maximum entropy approaches, particularly for unstructured objects like a line object (Fig.~\ref{Fig_gridworld_suboptimal}).

\par
To further quantitatively compare the performance of different approaches, four q-learning models were trained, separately, according to the three inferred reward maps and the ground truth. The number of training episodes was set to $500$, and the maximum number of steps for each episode was $100$. Then, $20$ points were randomly initialized in the grid world, and the four trained models were used to guide the agents to move towards the target position separately. Considering time efficiency, the trials were only considered successful when the agent can stop at the desired position within $100$ steps. The whole process was repeated $20$ times with random target positions. The final success rate over $400$ trials ($20 \times 20$) for different approaches is summarized in TABLE~\ref{Table_success_rate}. 

\par
It can be seen from TABLE~\ref{Table_success_rate} that the ME-IRL results in the worst case in our setup, where the success rate is only $9.3\%$ and $3.2\%$ for the point and line objects, respectively. In contrast, the result obtained using the ground truth is best ($99.3\%$ and $100\%$). The success rate achieved using the PTR and the proposed GPSR are $47.0\%$ and $72.0\%$ for the point object, respectively. Although GPSR results are still worse than the ground truth, the performance has already significantly improved from the state-of-the-art approaches ($62.7\%$ and $25\%$, respectively). In addition, regarding the line object, the success rate for PTR further decreases to $33.2\%$ while the GPSR increases the number to $84.5\%$. This is because PTR can only consider the individual demonstrations based on temporal information, while the proposed GPSR globally generates pair-wise comparisons between all available demonstrations using spatial information. But it needs to be noted that the presented comparison results are achieved when only a few demonstrations are available, and they are sub-optimal. If a large amount of demonstrations becomes available, the performance of PTR and ME-IRL could be improved. But, theoretically, they still lack the capability to successfully reconstruct the reward map of a ``line" object.
In addition, it is noteworthy that the MI-GPSR was not compared in the grid word experiments. This is because the state inputs (grid position) in the grid world are ideal without any noise. So there is no need to further disentangle the features in latent space.



\begin{table}[htbp]
\centering
 \caption{Performance of Learning from Demonstrations Methods}
 \label{Table_success_rate}
 \begin{tabular}{M{2.5cm}  M{2.0cm} M{2.0cm}}
  \toprule
  \multirow{2}*{\textbf{Methods}} & \multicolumn{2}{c}{\textbf{Success rate}}\\
                         & Point object & Line object\\
  \midrule
Ground truth & 99.3\%     & 100\% \\
ME-IRL       & 9.3\%      & 3.2\%\\
PTR          & 47.0\%     & 33.2\%\\
\textbf{GPSR}         & 72.0\%     & 84.5\%\\
  \bottomrule
 \end{tabular}
\end{table}

\subsection{Performance of the MI-GPSR on Unseen Demonstrations}
\par
\subsubsection{Training Details}\label{Sec_Phantom_training_details}

\par
Regarding the training process of the proposed MI-GPSR approach, only six demonstrations were randomly selected out of ten expert demonstrations. Five of them were used as training data set, while the remaining one was used for validation. To balance the weights of different demonstrations, all demonstrations were down-sampled to $100$ frames. Therefore, the size of the pair-wise training data set is $C(500, 2)=124,750$, while the size of the validation data set was $C(100, 2) = 4,950$. The network parameters of the proposed MI-GPSR are updated based on the three loss functions $\mathcal{L}_{rec}$, $\mathcal{L}_{MI}$, and $\mathcal{L}_{rank}^{s}$ in a collaborative fashion (see Algorithm~\ref{algorithm_network_update}). Other training details are defined as follows: batch size: $8$, epoch: $5$, leaning rate: $1\times10^{-5}$. In addition, all the training process was run on a single GPU (Nvidia TITAN Xp).



\begin{figure*}[ht!]
\centering
\includegraphics[width=0.98\textwidth]{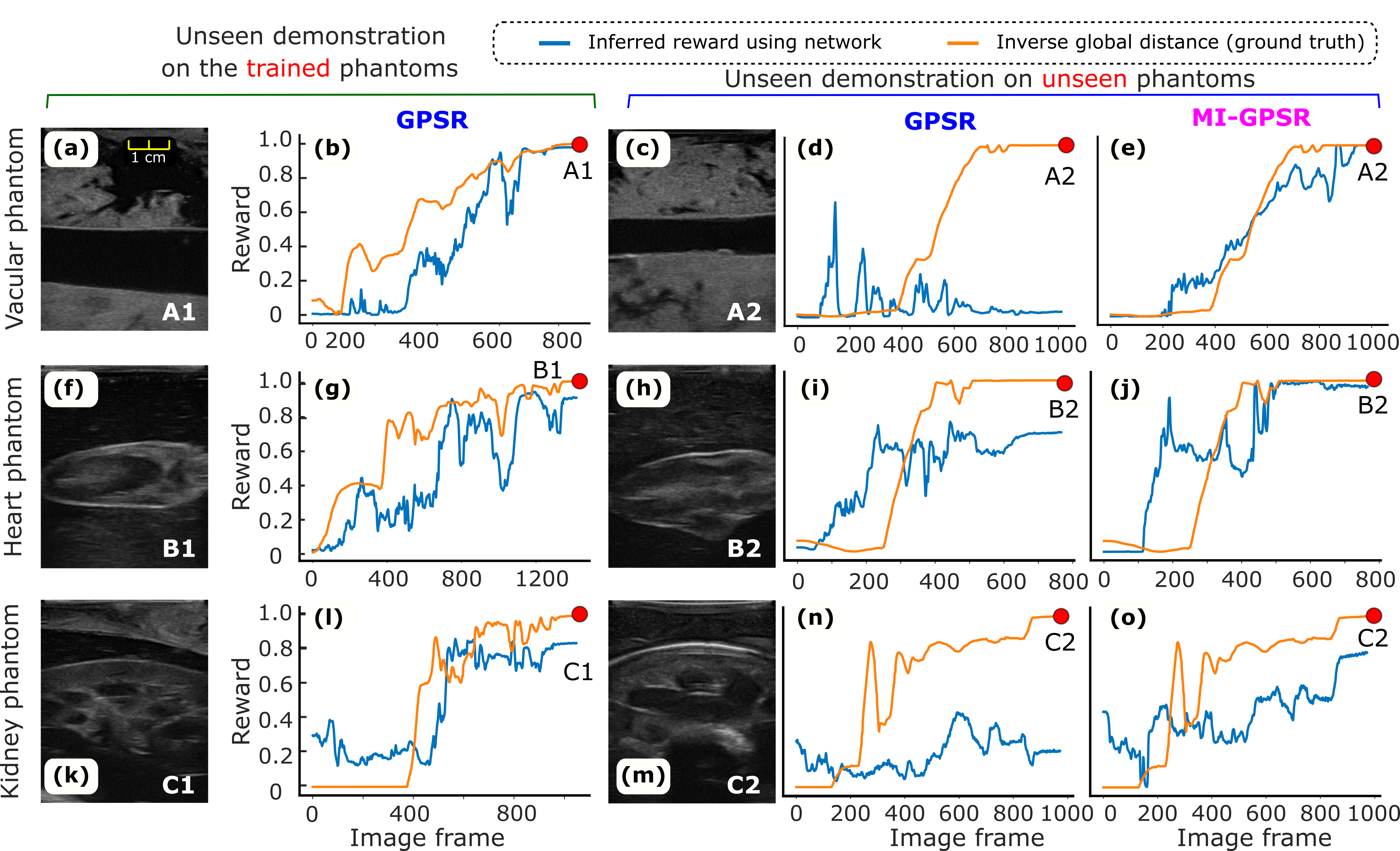}
\caption{Illustration of inferred rewards from three unseen US clips acquired from the trained phantoms, as well as three unseen demonstrations on three unseen phantoms (vascular phantom, ex-vivo chicken heart and lamb kidney phantom).
}
\label{Fig_rewad_curve}
\end{figure*}

\subsubsection{Inferred Reward from Unseen Demonstrations and Unseen Phantoms}
\label{Sec_Phantom_test}
\par
To validate the performance of the proposed models, we tested the well-trained GPSR model on unseen demonstrations acquired from the same phantoms as training data. Considering inevitable variations for different patients, we further tested the GPSR and MI-GPSR on the demonstrations acquired from unseen phantoms. As defined in Section~\ref{sec:III-GPSR}, the desired reward is negatively correlated to the global generalized distance $D_i^k$. For better visualization, the inverse global generalized distance ($1 -D_i^k$) is employed as a reference index, shown as orange curves in Fig.~\ref{Fig_rewad_curve}. The inferred reward of each frame by the proposed networks (GPSR and MI-GPSR) are depicted as the blue curve in Fig.~\ref{Fig_rewad_curve}. The reward is estimated purely based on US images without any additional information during the inference process.

In Fig.~\ref{Fig_rewad_curve}, the images in the first column are the last frames (A1, B1 and C1) of the unseen demonstrations obtained on the vascular phantom, heart phantom, and lamb kidney phantom, respectively. The inferred rewards of the unseen demonstrations by GPSR are depicted in Fig.~\ref{Fig_rewad_curve}~(b), (g) and (l). It can be seen from the figure that the change tendency of the predicted rewards are consistent with the reference index. For the vascular phantom, the predicted reward of the final frame of the demonstration achieved the highest reward value ($0.98$). For the more changeling task of ex-vivo phantoms (chicken heart and lamb heart), the inferred rewards of the last frames also achieve $0.90$ and $0.84$, respectively.

\par
The aforementioned results are achieved on the same phantoms as the training data. To mimic the variations between patients, three new phantoms with the same anatomies are built respectively. Comparing the images presented in Fig.~\ref{Fig_rewad_curve} (c), (h), and (m) with the ones listed in the first column, the sizes, shapes of the involved tissues, and artifacts of the image are different. The same expert is asked to perform three demonstrations on each unseen phantom, respectively. Due to the significant difference, the GPSR fails to predict the reward for the vascular and kidney phantoms ($0.01$ and $0.21$ for the last frame). The GPSR inference result of the heart phantom of the last frame achieved $0.70$, while it is still significantly less than the one achieved on the trained phantom ($0.92$). We consider the relatively high reward inference of the heart phantom mainly because the new phantom is still similar to the one used for training (see Fig.~\ref{Fig_rewad_curve} (f) and (h)).

\par
To improve the generalization capability of the reward prediction network to tackle inter-patient variations, we further proposed MI-GPSR to explicitly separate the task-related and domain features of input images. Then, only the task-related features are used to compute the reward. The MI-GPSR results on the same demonstrations acquired from the unseen phantoms are depicted in Fig.~\ref{Fig_rewad_curve} (e), (j), and (o), respectively. In all cases, MI-GPSR achieved much better results than the ones obtained by GPSR. For the unseen phantom, the predicted rewards of the last frame in the demonstrations by the MI-GPSR are $1.0$ and $0.98$ for the vascular and heart phantom, respectively. Similar to the second column, the predicted reward of the kidney phantom is slightly lower than others ($0.78$). We consider that this phenomenon is mainly caused by the more complex structures of the kidney in the B-mode images. To further improve its performance on complex anatomy as well, more data can be recorded from the expert.

\subsection{\revision{Performance of the MI-GPSR on Unseen Demonstrations from Volunteers}}
\par
\revision{In order to ensure the transferability of the proposed MI-GPSR in real scenarios, we further compared its performance with GPSR on in-vivo carotid data recorded from six healthy volunteers. To validate the generalization capability on different patients, six demonstrations (two for each volunteer) recorded from three volunteers' carotids were used as the training dataset, while three distinct demonstrations (one for each volunteer) recorded from three unseen volunteers were used for testing. The expert demonstrations were given by manually maneuvering the linear probe (SIEMENS AG, Germany) fixed at the robotic end-effector. The acquisitions were performed within the Institutional Review Board Approval by the Ethical Commission of the Technical University of Munich (reference number 244/19 S). A certified Siemens Juniper US Machine (ACUSON Juniper, SIEMENS AG, Germany) was used with the default US parameters for carotid imaging to acquire in-vivo data. The US images from the Juniper US machine were accessed in real-time using a frame grabber (MAGEWELL, Nanjing, China). It is worth noting that both GPSR and MI-GPSR models were retrained for human carotid scans. The training details were the same as the ones used for phantoms (see Sec.~\ref{Sec_Phantom_training_details}).}


\begin{figure*}[ht!]
\centering
\includegraphics[width=0.90\textwidth]{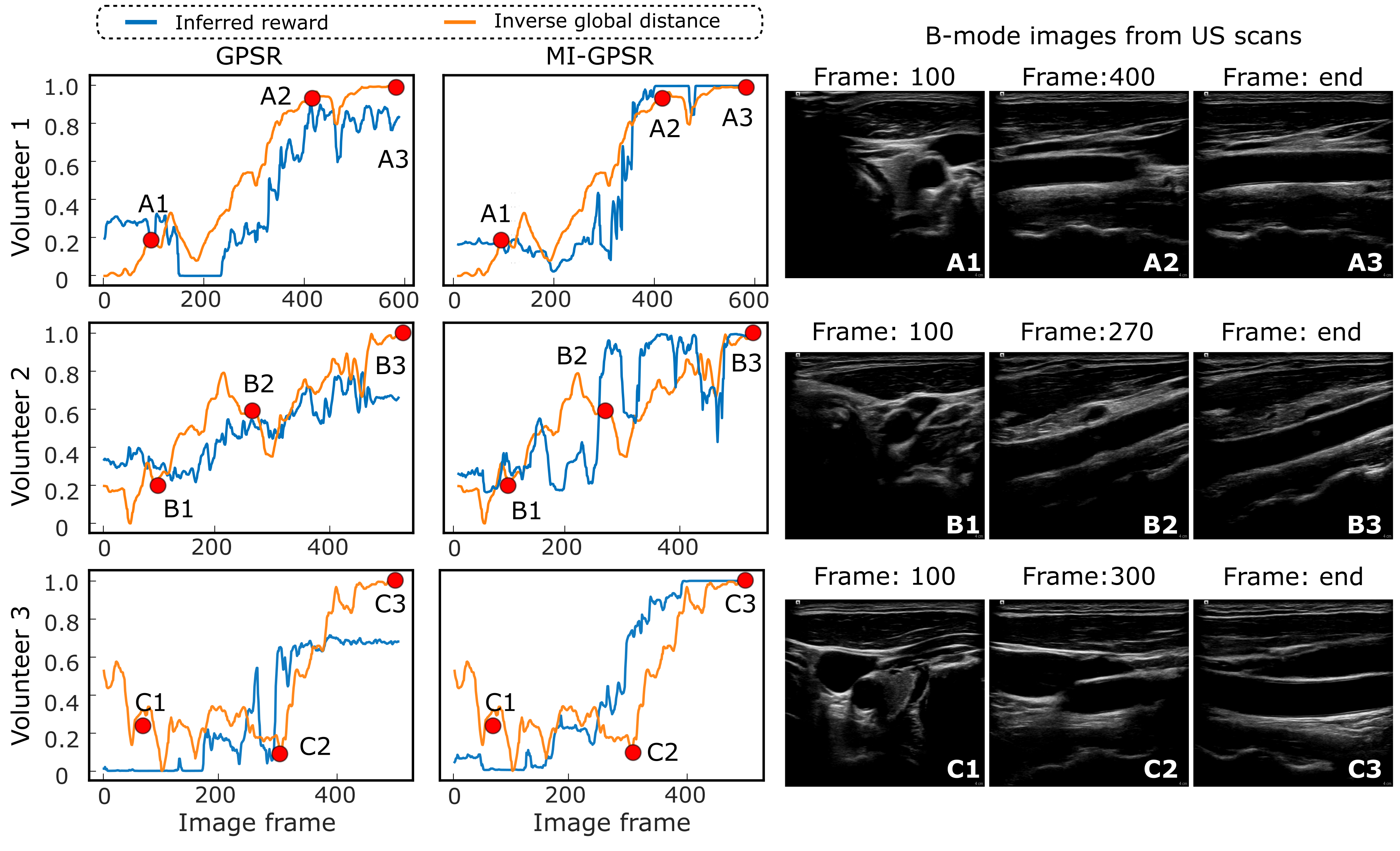}
\caption{\revision{Illustration of inferred rewards computed using GPSR and MI-GPSR from three unseen US carotid clips acquired from three unseen volunteers. The three volunteers' data was not involved in the training data at all. Three rows represent the results of the unseen clips recorded on three unseen volunteers.}
}
\label{Fig_human_rewad_curve}
\end{figure*}

\par
The interpreted rewards for individual images from unseen demonstrations on unseen volunteers have been depicted in Fig.~\ref{Fig_human_rewad_curve}. The reward was computed purely based on US images without additional information during inference. Each row represents the results of one volunteer. To provide an intuitive understanding of in-vivo data, two intermediate images and the last frames in each demonstration have also been shown in Fig.~\ref{Fig_human_rewad_curve}. It can be seen from Fig.~\ref{Fig_human_rewad_curve} that GPSR can roughly track the changes in the inverse global distance (orange curves). However, the interpreted reward ($\in[0,1]$) curves using GPSR are not as accurate as the ones computed using MI-GPSR. More fluctuations are witnessed in the end phase for all three unseen demonstrations, which will hinder the achievement of an accurate understanding of the sonographer's intention. The predicted reward values of the three last frames (A3, B3, and C3) of the three unseen volunteers are $0.82$, $0.65$, and $0.67$, respectively. In contrast, the reward values of the same three frames, namely the target standard plane, are accurately predicted by the evolved MI-GPSR as $0.99$, $0.99$, and $0.99$, respectively. Regarding the intermediate frames B2, the computed reward using GPSR and MI-GPSR are $0.53$ and $0.95$, respectively. By comparing the images in B2 frame, it is quite similar to the standard plane shown in the last frame (B3) for Volunteer 2. Thereby, we consider the MI-GPSR can better predict the reward of individual B-mode images.

\par
The proposed MI-GPSR can significantly improve the generalization capability to adapt patient-specific data with a limited training dataset. This is consistent with the results on three types of vascular and ex-vivo phantoms. By comparing the results on unseen volunteers (Fig.~\ref{Fig_human_rewad_curve}) to the ones obtained on unseen phantoms (Fig.~\ref{Fig_rewad_curve}), we can find GPSR performs much better on human carotids than on phantoms. This is different from what we thought before the experiments. We consider the reason for this phenomenon to be that human tissues (same organs) usually share much more \final{structural} similarities than phantoms, particularly when a generally optimal US acquisition setting is used. 
\final{It can be observed from Fig.~\ref{Fig_human_rewad_curve} that although the carotid images (each row) are obtained from three different volunteers, the structural information, including background intensity and surrounding tissues, is similar. In contrast, the custom-made phantom was made by randomly mixing paper pulp into the gelatin power. It is evident from Fig.~\ref{Fig_rewad_curve} that the background information in the training and testing phantoms differs significantly from each other. Consequently, better performance is observed with using GPSR on human carotid data compared to the phantom data.}

\subsection{Performance of Standard Scan Planes Alignment}
In order to validate the proposed approach in realistic cases, the mimicked vascular phantoms with a ``line" objective and ex-vivo animal organ phantoms with a ``point" objective were built. To guarantee the generalization capability, the neural MI estimator is employed to explicitly disentangle the task-related and domain features in the latent space. To validate whether the proposed method can robustly deal with an unseen case, two mimicked tubular structures located at different depths (vessel $1$: $36~mm$ versus vessel $2$: $27~mm$) of the custom-designed phantoms were used. All the training data was recorded from the vessel $1$. To further increase the differences (in both geometry and style) between the B-mode images obtained from the two phantoms, a different brightness $48~dB$ was used for the unseen vessel $2$ from the default one ($66~dB$) used in other acquisitions. Two significantly different images obtained from vessel $1$ and vessel $2$* (a different acquisition setting) are depicted in Fig.~\ref{Fig_human_comparision}. The autonomous navigation results obtained using GPSR and MI-GPSR on the unseen vessel $2$* are described as well.
Besides, to compare the performance between the proposed approach and a human expert, the variations on final position and orientation for different experiments were recorded. Finally, considering that objects could be moved after training, the phantoms were rotated into different angles ($30,~60,~\text{and}~90^{\circ}$) to validate whether the method was still able to place the probe in the target planes correctly.

\subsubsection{Comparison with a Human Expert and the Validation of the Generalization Capability}
\par
To quantitatively compare the performance between an experienced human expert, GPSR and MI-GPSR approaches, ten navigation experiments were performed by the expert and the proposed GPSR approach, respectively, to position the probe along the longitudinal plane of the vessel $1$. The mimicked blood vessels are straight holes inside the phantom, allowing positional slack in the direction of the vessel centerline. Thus, the distance error ($e_d$) is defined as the distance between the vessel centerline and the US imaging plane in 3D space instead of the absolute positional difference. The centroid of the vessel in US images was calculated using the same steps as~\citep{jiang2021autonomous}: (1) using a U-net to segment the vessel from US images, (2) computing the centroid based on the binary map using OpenCV, and (3) applying the spatial calibration result to calculate the 3D position of pixel-wise centroid in 2D images. The ground truth of the vessel centerline was computed by moving the probe along the vessel centerline. Besides metric $e_d$, the absolute rotation error ($e_r$) between the probe long axis ($^{p}X$) and the vessel centerline was further defined to assess the probe orientation accuracy.


\par
\begin{figure}[ht!]
\centering
\includegraphics[width=0.45\textwidth]{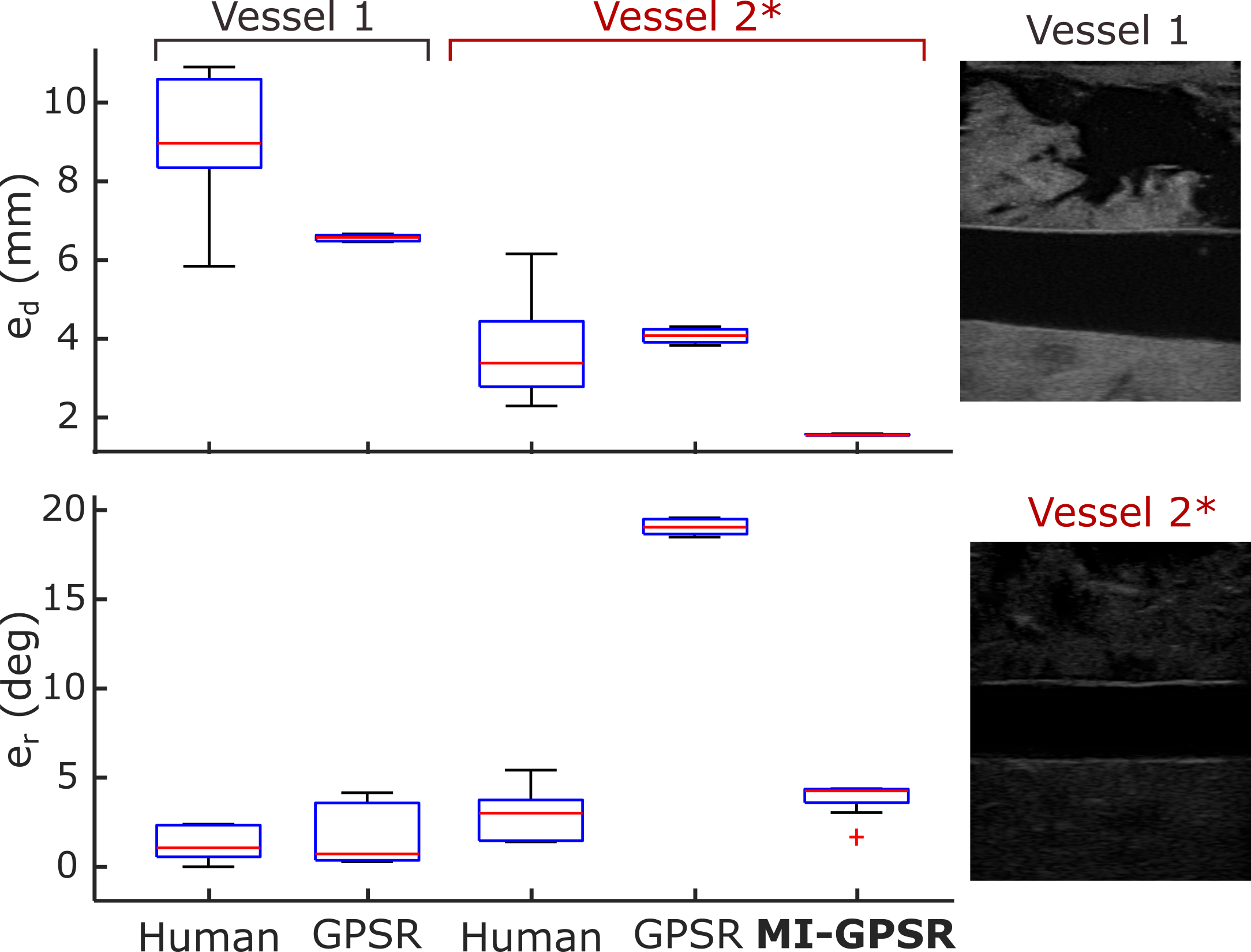}
\caption{The performance of an experienced human expert, GPSR and MI-GPSR on two different vessels. All the training data are only obtained from vessel 1. To validate the generalization capability of the proposed MI-GPSR, an unseen vessel 2 with a smaller diameter is employed. The sign ``*" means the US brightness value is $48~dB$ in acquisitions, which is different from the default one ($66~dB$) used in other acquisitions. 
}
\label{Fig_human_comparision}
\end{figure}

The results of $e_d$ and $e_r$ over the $20$ trials (ten for each phantom) performed by the same expert and the GPSR approach on vessel $1$ are summarised in Fig.~\ref{Fig_human_comparision}. Regarding $e_d$, GPSR results are more concentrated and the average $e_d$ (Mean$\pm$STD: $6.6\pm 0.1~mm$) is less than the one obtained from the expert ($9.1\pm 1.7~mm$). Regarding $e_r$, the GPSR achieves comparable results ($1.5\pm 1.6^{\circ}$) to the expert ($1.3\pm 1.0^{\circ}$). Based on a significant test ($t$-test), the probability between $e_r$ obtained from the expert and the GPSR is $0.72>0.05$, which means there is no significant statistical difference between these data.

\par
To further compare the generalization capability of the GPSR and the evolved MI-GPSR, an unseen blood vessel, namely vessel $2$, was employed to represent a patient-specified anatomical structure. To perform such validations in a challenging case, a different brightness of $48~dB$ was used for unseen vessel $2$ rather than the default one ($66~dB$) used in other acquisitions. The resulting images obtained from vessel $2$* become significantly different from the ones used in training from vessel $1$. 
To compare the human performance for different vessels, ten independent trials were carried out by the same human expert. Without further updating the parameters in the trained reward network using the demonstrations for vessel $1$, the GPSR and the evolved MI-GPSR was also run ten times, separately, on unseen vessel $2$*. The navigating results are depicted in Fig.~\ref{Fig_human_comparision}. 

\par
In terms of $e_r$, the expert achieved close results on two different phantoms ($1.5\pm 1.6^{\circ}$ and $2.9\pm 1.3^{\circ}$) while $e_d$ obtained for vessel $2$* is significantly smaller than the one obtained on vessel $1$ ($3.7\pm 1.1~mm$ and $9.1\pm 1.7~mm$). We consider the observed intra-operator variations of a positional error on different phantoms are caused by the geometry characteristic of tubular structures. Due to the limited human spatial perception, experts determine the standard US plane (longitudinal view) only based on B-mode images rather than probe position. Considering the tubular structures, the target geometry in the resulting 2D images will be changed significantly from an ellipse to a rectangle when the probe rotates around the probe centerline; in contrast, the translational deviation orthogonal to the vessel centerline will only change the height of the resulting rectangle shape (longitudinal view). The significant changes in terms of shape are beneficial for human experts to capture the rotational variations, while the translational motion-induced rectangle height change is not easily visible for humans. Therefore, we think biased $e_d$ results for the same expert on two different phantoms could happen, particularly for the tubular phantoms. 

\begin{figure*}[ht!]
\centering
\includegraphics[width=0.85\textwidth]{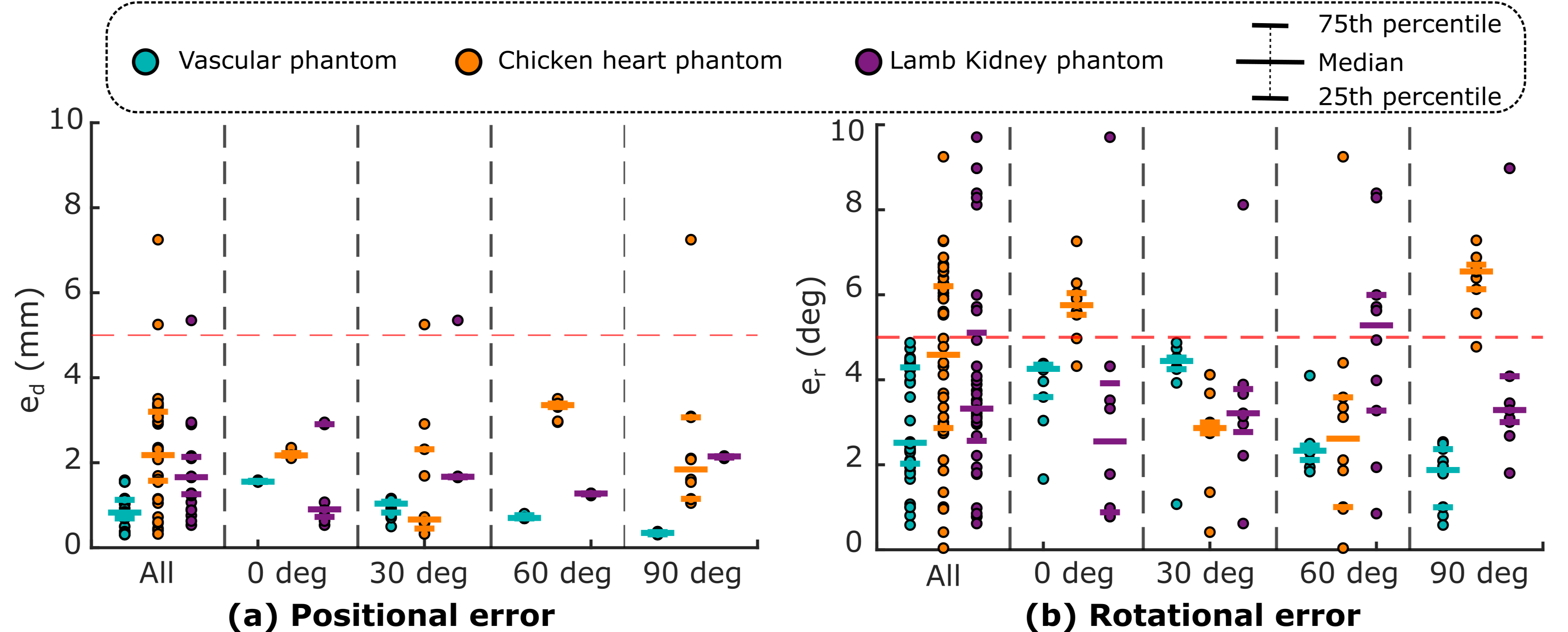}
\caption{The spatial alignment errors between the target probe pose and the automatically identified pose using the proposed MI-GPSR approach.}
\label{Fig_result_all_phantoms}
\end{figure*}

\par
It can be seen from Fig.~\ref{Fig_human_comparision}, $e_r$ results of the GPSR and MI-GPSR are $19.1\pm0.4^{\circ}$ and $3.8\pm0.9^{\circ}$, respectively. Based on this, we can conclude that the GPSR failed on significantly different vessel $2$*, while the proposed MI-GPSR can still accurately navigate the probe to the standard planes in terms of angular error.The results obtained by MI-GPSR are comparable to the ones obtained by the human operator ($3.8\pm0.9^{\circ}$ vs $3.7\pm 1.1~mm$). However, it is noteworthy that although the $e_r$ is very large, the positional errors of GPSR are still comparable to the human operators, even more accurate and concentrated ($1.3\pm0.4~mm$ vs $2.9\pm 1.3^{\circ}$). We consider this because that human experts usually can quickly move the probe toward the target vessel and then slowly rotate the probe to find the objective plane. During the rotational adjustment, the probe tip position is not changing. Such behavior is implicitly represented in the expert demonstrations and can be learned by the GPSR (also for the evolved MI-GPSR) by generating pairwise image comparisons. The $e_r$ of the proposed MI-GPSR is only $1.5\pm0.1~mm$, which is more stable than human experts. 
These results demonstrate that the MI-GPSR has the potential to deal with inter-patient variations occurring in real scenarios. In addition, it is also noteworthy that the MI-GPSR can generate better results than individual demonstrations, i.e., $e_d$. This is because the examples with low confidence values have been filtered out and also the demonstrations are considered globally to balance the variation between individual demonstrations.

\subsubsection{Quantitatively Assessment of Standard Planes Alignment on Various Anatomical Targets}
\par
This section demonstrates the quantitative results of the proposed MI-GPSR approach on different objects (tubular tissues, chicken heart, and lamb kidney phantoms). It is notable that the phantom used for testing here is different from the phantom for training. To validate whether the MI-GPSR has the potential to be used in real scenarios, the phantoms are placed in different positions and orientations during the experiments. This is because patients' positions in different examinations are different. For each phantom, four groups of experiments with different angles ($0, 30, 60,~\text{and}~90^{\circ}$) rotated around a given point on the flat table were carried out. Each group consisted of ten independent experiments. In each experiment, the 3D scan is needed to compute the coarse localization in the current setup; then followed by a fine-tuning process to overcome the potential variations between simulated images and real images.

\par
The performance of the final alignment was assessed using positional ($e_d$) and rotational ($e_r$) errors as defined in the last subsection. However, differing from the $e_d$ defined for vascular phantom, the absolute positional error, computed between the final position achieved using the MI-GPSR approach and the ground truth, was used for kidney and chicken heart phantom. This is because that the standard planes are only able to be obtained by a unique pose for the ``point" objective. The ground truth of each case was manually determined by the mean of the final frames' poses in ten expert demonstrations. The final performance is summarized in Fig.~\ref{Fig_result_all_phantoms}. 

\par
\final{In general, $e_d$ for tubular structure, chicken heart and lamb kidney are $0.9\pm 0.4~mm$, $2.4\pm1.3~mm$ and $1.7\pm0.9~mm$, respectively. Besides, $e_r$ for the three different phantoms are $3.0\pm1.3^{\circ}$, $4.4\pm2.2^{\circ}$, and $4.0\pm2.5^{\circ}$, respectively.} Compared to the results obtained on ex-vivo phantoms, the average $e_d$ and $e_r$ obtained from the vascular phantom are slightly smaller and more concentrated. 
We consider the B-mode image obtained from the custom-built vascular phantom is more stable and closer to the training data than the ones obtained from the animal organs. Thereby, the slight large variation caused by practical factors like the force-induced deformation results in relatively larger navigation errors for the chicken heart and lamb kidney phantoms.


\par
It can be seen from Fig.~\ref{Fig_result_all_phantoms} that the trained MI-GPSR model can also robustly work when the phantoms are rotated in different angular deviations without the requirement for further training. In the worst case, the maximum $e_d$ and $e_r$ for all phantoms are $7.2~mm$ and $9.7^{\circ}$, while the best results can achieve $0.3~mm$ and $0.04^{\circ}$, respectively. Overall, both positional and rotational errors are mainly distributed below $5~mm$ and $5^{\circ}$, respectively, among all experiments carried out on different phantoms. This means that the proposed approach can automatically identify the standard planes for challenging ex-vivo animal organs even when the object's position and orientation change. This capability enables the possibility of addressing the practical factors, i.e., patients' position variations, for different trials in real scenarios.

\par
Besides the navigation accuracy, the time periods required for the main components of the proposed MI-GPSR approach are also summarized as follows: the simulation process costs $244~s$ for generating $4,643$ images ($52~ms$ for each image), and the reward network costs $11\pm4~ms$ for computing the reward for each image. In terms of time efficiency, the proposed approach is more practical than the RL-based methods~\citep{li2021autonomous,hase2020ultrasound} in real scenarios because it can significantly reduce the dynamic interaction with patients. 
\section{Discussion}
\par
Considering the potential inter-operator variation of free-hand US examinations, only one experienced expert was employed to both give the training demonstrations and obtain experimental results in this work. In such a way, we consider a fair comparison can be carried out between the human expert and the proposed MI-GPSR approach. Technically, the proposed MI-GPSR can learn operation skills from multiple experts as well if there is no significant inter-operator variation witnessed in the given demonstrations. \final{While we emphasize in the article that MI-GPSR is introduced to tackle the issue of sub-optimal demonstrations, it's worth noting that the framework can also be applied to optimal US demonstrations. If the demonstrations can terminate at the desired US standard planes, both GPSR and MI-GPSR can be trained to learn the implicit reward function using the self-supervised probabilistic ranking approach.}

\par
The MI metric was employed to explicitly disentangle the task-related features and domain features. By removing variable domain features caused by patient-specific variations, the generalization capability of MI-GPSR can be enhanced.
\revision{However, it is noteworthy that the whole framework (learning from demonstration and autonomous navigation) was only validated on different phantoms in the current study.} Considering the significant variations among human tissue properties, the reproducibility of final probe navigation results could be reduced in real scenarios. To tackle this issue, either a post-processing approach can be used to correct the image deformation after scans~\citep{sun2010trajectory,jiang2021deformation,jiang2023defcor} or an online tuning process of the contact force could be implemented based on real-time feedback~\citep{virga2016automatic}. Both types of approaches require a deep understanding of US images, while there is no unified quality assessment metric yet, and the definition of quality could be different for various applications. Thereby, future work should further clearly define the definition of imaging quality for given downstream applications, which will further enable the automatic contact force optimization procedure. \revision{In addition, to bridge the gap between the phantom trials and real patients, some practical factors should also be properly considered, e.g., patient physiological movements during the scans. The research community has already noted these problems, and there are some emerging articles focusing on monitoring and compensating for rigid motion~\citep{jiang2021motion, jiang2022precise}, articulated motion~\citep{jiang2022towards}, and breathing motion~\citep{dai2021deep}. }

\par
\revision{
In addition to the primary objective of developing an autonomous RUSS, we believe understanding the ``language of sonography" from expert demonstrations could be as valuable as the advancements made in the robotic US examination itself. To ensure the proposed MI-GPSR can also effectively recover the ``language of sonography" for real human examination, we recorded examination demonstrations for carotid on six volunteers. The results obtained on unseen demonstrations from unseen volunteers demonstrate the proposed MI-GPSR can precisely predict the reward for individual images. 
\final{To avoid misleading information, we want to emphasize that specific anatomical images recorded from unseen volunteers still share the same shape priors and statistical distributions as the training data. In order to extend to unseen anatomies, new images should be obtained to retrain or refine the network models.}
To inspire further study, it is also worth noting that the baseline GPSR achieved better results on unseen demonstrations of human carotid (Fig.~\ref{Fig_human_rewad_curve}) than of unseen phantoms (Fig.~\ref{Fig_rewad_curve}). We consider this is because that human tissues (same organs) usually share much more similarities than homemade phantoms with randomly mimicked noise backgrounds. This finding suggests that the collection of massive examination demonstrations from real patients holds good potential to contribute to the understanding of the common knowledge underlying US examinations. The research in the direction of recovering the ``language of sonography" will benefit in transferring senior sonographers’ physiological knowledge and experience to novices.}

\section{Conclusion}
In this work, we present an advanced machine learning framework to automatically discover standard planes based on a limited number of expert demonstrations. \final{To understand the ``language of sonography", a neural reward network is optimized to compute reward value for individual images based on the given expert's demonstrations.} Besides, to guarantee the generalization capability to overcome the inter-patient variations, the proposed MI-GPSR explicitly disentangles the task-related and domain features in the latent space. To validate the proposed approach, experiments were performed on typical ``line" targets (vascular phantoms) and ``point" objects (ex-vivo chicken heart and lamb kidney phantoms), respectively. The experimental results of the MI-GPSR on tubular structure, chicken heart and lamb kidney phantoms are $0.9\pm 0.4~mm$, $2.4\pm1.3~mm$ and $1.7\pm0.9~mm$ in terms of $e_d$, and are $3.0\pm1.3^{\circ}$, $4.4\pm2.2^{\circ}$, and $4.0\pm2.5^{\circ}$ in terms of $e_r$, respectively. \revision{To show the potential to be used on real images from humans, the performance of reward inference was also validated on the carotid images recorded from six volunteers.}
Several conclusions can be drawn from this study: 1) Learning high-level physiological knowledge from human experts from demonstrations is feasible using a reward network; 2) The explicit disentanglement process of latent space features can significantly contribute to improving the generalization capability of the proposed MI-GPSR to overcome the variation between the trained data and unseen phantoms.

\par
Beyond the contributions to autonomous robotic US navigation, we also want to emphasize the importance of the side results of this study about the understanding and modeling of expert sonographers' semantic reasoning and intention. We believe that the understanding of the ``language of sonography" could be considered as valuable and essential as the progress made in robotic sonography itself. This can not only allow for autonomous intelligent RUSS development but also for designing US education and training systems and advanced methodologies for grading and evaluating the performance of human and robotic sonography. In the future, the proposed intelligent robotic sonographer can be extended by further considering the practical factors in real scenarios and testing the method on real patients. Such an intelligent system will have the potential to enable the development of a fully automatic intervention system (i.e., vessels~\citep{chen2020deep}) and extensive US examination programs for the early diagnosis and monitoring of internal lesions or tumors.

\section*{ACKNOWLEDGMENT}
The authors would like to thank Dr. Med. Reza Ghotbi from the vascular surgery department of Helios Klinikum München West and Dr. Med.  Angelos Karlas from the vascular surgery department of Klinikum Rechts der Isar for their valuable feedback and insightful discussions. \revision{In addition, the authors would like to acknowledge the Editors and volunteering reviewers for their time and implicit contributions to the improvement of the article's thoroughness, readability, and clarity.}

\bibliographystyle{SageH}
\bibliography{references}

%






\end{document}